\documentclass[runningheads]{llncs}

\usepackage[mobile]{eccv}

\usepackage{eccvabbrv}

\usepackage{graphicx}
\usepackage{booktabs}

\usepackage[accsupp]{axessibility}  

\usepackage{hyperref}

\usepackage{orcidlink}

\usepackage{multirow}
\usepackage{multicol}
\usepackage{amssymb}
\usepackage{xcolor}
\usepackage{colortbl}
\usepackage{wrapfig} 
\newcommand{\cmark}{\textcolor{green}{\checkmark}}
\newcommand{\xmark}{\textcolor{red}{\textsf{X}}}

\newcommand{\ourmodel}{DA-Flow}
\definecolor{best}{HTML}{BDDDD5}
\definecolor{secondbest}{HTML}{FFF3B3}

\begin{document}

\title{DA-Flow: Degradation-Aware Optical Flow Estimation with Diffusion Models} 

\titlerunning{Degradation-Aware Optical Flow Estimation with Diffusion Models}

\author{Jaewon Min\inst{1} \and
        Jaeeun Lee\inst{1} \and
        Yeji Choi\inst{1} \and 
        Paul Hyunbin Cho\inst{1} \and 
        Jin Hyeon Kim\inst{1} \and  
        Tae-Young Lee\inst{2} \and
        Jongsik Ahn\inst{2} \and 
        Hwayeong Lee\inst{2} \and
        Seonghyun Park\inst{2} \and
        Seungryong Kim\inst{1}$^\dagger$
        } 
\authorrunning{J. Min et al.}

\begingroup
\renewcommand{\thefootnote}{}
\footnotetext{$^\dagger$: Corresponding author}
\endgroup

\institute{$^{1}$KAIST AI \quad  $^{2}$Hanwha Systems\\
\textcolor{Blue}{\href{https://cvlab-kaist.github.io/DA-Flow}{https://cvlab-kaist.github.io/DA-Flow}}}

\maketitle

\begin{abstract}
    Optical flow models trained on high-quality data often degrade severely when confronted with real-world corruptions such as blur, noise, and compression artifacts. To overcome this limitation, we formulate \textbf{Degradation-Aware Optical Flow}, a new task targeting accurate dense correspondence estimation from real-world corrupted videos. Our key insight is that the intermediate representations of image restoration diffusion models are inherently corruption-aware but lack temporal awareness. To address this limitation, we lift the model to attend across adjacent frames via full spatio-temporal attention, and empirically demonstrate that the resulting features exhibit zero-shot correspondence capabilities. Based on this finding, we present \textbf{\ourmodel}, a hybrid architecture that fuses these diffusion features with convolutional features within an iterative refinement framework. \ourmodel\ substantially outperforms existing optical flow methods under severe degradation across multiple benchmarks.
\end{abstract}    

\section{Introduction}
\label{sec:intro}
Optical flow estimation, the task of estimating per-pixel motion fields between consecutive video frames, is a fundamental dense correspondence problem in computer vision. With the advent of deep neural networks~\cite{teed2020raft, wang2024sea, poggi2025flowseek}, recent methods have achieved remarkable accuracy. 

Real-world videos are rarely clean; motion blur, sensor noise, compression artifacts, and low resolution frequently co-exist, severely degrading visual quality. Despite the prevalence of such degradations, how optical flow models behave under such degradations remains largely unexplored. 
Recently, RobustSpring~\cite{schmalfuss2025robustspring} provided the first comprehensive study on the robustness of dense matching models, benchmarking their generalization from clean synthetic training data to a wide spectrum of real-world degradations.
Despite this systematic analysis, a fundamental question remains open: \textbf{is it truly impossible to accurately estimate optical flow from corrupted inputs?} 

\begin{figure}[t]
    \centering
    \includegraphics[width=1\textwidth]{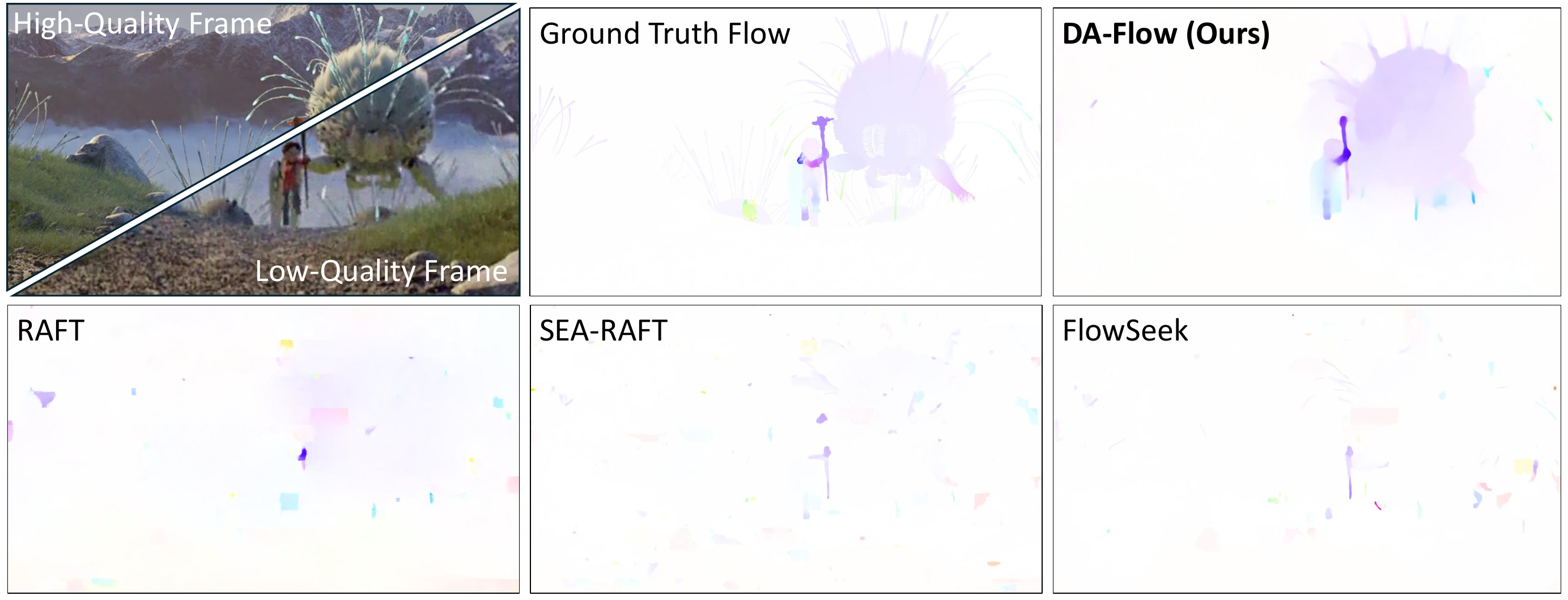}
    \caption{\textbf{Qualitative comparison of \ourmodel\ with baselines~\cite{teed2020raft, wang2024sea, poggi2025flowseek} on Spring benchmark~\cite{Mehl2023_Spring}}. Under severe degradations, existing optical flow methods fail to recover reliable correspondences, whereas our \ourmodel\ accurately estimates the underlying motion.}
    \label{figs:teaser}\vspace{-10pt}
\end{figure}

Motivated by this question, we shift the focus from robustness to accuracy by introducing a new task, \textbf{Degradation-Aware Optical Flow}, that directly estimates dense correspondences from severely degraded inputs. This task is fundamentally ill-posed: degradations destroy fine textures and attenuate motion boundaries, leaving insufficient visual evidence for reliable matching. In such regimes, correspondence estimation is not merely a matter of distribution shift but becomes inherently ambiguous. Simply augmenting clean training data with synthetic corruptions does not adequately address this challenge; what is needed are representations that are both rich enough to preserve spatial structure for dense matching and sensitive to degradation patterns to recover information lost during corruption, as illustrated in Fig.~\ref{figs:teaser}.

Recent works~\cite{nam2023diffusion, zhang2023tale, tang2023emergent, nam2025emergent, ke2024repurposing, ke2025marigold, kim2025seg4diff, tian2024diffuse} have shown that intermediate features of diffusion models encode rich structural and semantic information, achieving strong performance on correspondence tasks~\cite{nam2023diffusion, zhang2023tale, tang2023emergent, nam2025emergent} as well as diverse downstream vision tasks such as depth estimation~\cite{ke2024repurposing, ke2025marigold} and segmentation~\cite{kim2025seg4diff, tian2024diffuse}. These findings suggest that diffusion representations capture geometric and structural cues far beyond what is needed for generation alone. Building on this insight, we observe that diffusion models trained for image restoration~\cite{lin2024diffbir, stablesr, ai2024dreamclear, duan2025dit4sr} offer an even more suitable foundation. Image restoration is likewise a highly underdetermined inverse problem, and models trained for this task must learn to recover clean structures from degraded inputs. As a result, their intermediate features naturally encode degradation patterns while preserving underlying scene geometry. This motivates our core design choice: leveraging restoration diffusion features to obtain representations that are degradation-aware, structurally rich for dense matching, and equipped with generative priors that can reason beyond corrupted observations. However, these features lack temporal awareness, limiting their effectiveness in producing accurate features for optical flow estimation.

Since our task involves video, a natural consideration is whether video restoration diffusion models~\cite{xie2025star, chen2025doveefficientonestepdiffusion, zhuang2025flashvsr}, which jointly model degradation and temporal dynamics, could serve as the backbone. However, such models often encode a stack of degraded frames into a temporally compressed latent representation through 3D convolutions or temporal attention. This produces a shared latent tensor where the temporal axis is entangled early in the encoding pipeline. While this design suits perceptual video restoration, where temporal smoothness and global consistency are desirable, it is structurally misaligned with dense correspondence estimation. Optical flow requires comparing spatial features extracted independently from each frame to establish pixel-level correspondences. When degraded frames are jointly encoded into a shared spatio-temporal latent space, their per-frame spatial structure is no longer preserved as separable entities, making the representation ill-suited for explicit pairwise feature matching.

To reconcile degradation-aware representation with the structural requirements of dense matching, we take a different approach: instead of adopting a monolithic video diffusion backbone, we start from a pretrained image restoration diffusion model~\cite{duan2025dit4sr} that preserves full spatial resolution at the frame level. We then lift it to handle multiple frames by injecting cross-frame attention across all layers. This design maintains independent spatial latents for each frame, which is crucial for dense matching, while enabling controlled temporal interaction for motion reasoning. By inheriting strong degradation-aware priors from image restoration pretraining and avoiding temporal latent collapse, our architecture yields representations intrinsically suited for dense correspondence estimation under severe corruption, while remaining substantially more efficient than video diffusion architectures.

Building on this representation, we introduce \textbf{\ourmodel}, a \textbf{D}egradation-\textbf{A}ware Optical \textbf{Flow} network built on top of RAFT~\cite{teed2020raft}. As illustrated in Fig.~\ref{figs:main_figure}, \ourmodel\ combines upsampled diffusion features from the lifted model with conventional CNN-based encoder features into a hybrid representation, enabling the correlation and iterative update stages to benefit from both degradation-aware structural priors and fine-grained spatial detail. Since ground-truth optical flow for real-world degraded videos is unavailable, we train \ourmodel\ using pseudo ground-truth flow generated by applying a pretrained flow model~\cite{wang2024sea} to a high-quality video, while feeding the corresponding degraded frames as input. We evaluate on degraded versions of established optical flow benchmarks~\cite{Mehl2023_Spring, butler2012naturalistic, wang2020tartanair} constructed via realistic degradation pipelines~\cite{chan2022investigating, wang2021real}, and demonstrate that \ourmodel\ achieves accurate flow estimation even under severe corruption where existing methods fail.

Our main contributions are as follows:
\begin{itemize}
    \item We formulate \textbf{Degradation-Aware Optical Flow}, a new task that estimates accurate dense correspondences from severely corrupted videos.
    \item We lift a pretrained image restoration diffusion model by introducing inter-frame attention and verify that its features encode geometric correspondence even under severe corruption.
    \item We introduce \textbf{\ourmodel}, a degradation-aware optical flow network that substantially outperforms existing methods on degraded inputs.
\end{itemize}
\section{Related Work}
\label{sec:related}
\subsubsection{Optical flow estimation.}
Optical flow estimation aims to model dense pixel-level motion between consecutive frames and serves as a fundamental component in various video-related tasks, including video generation and scene reconstruction. Modern deep learning approaches have significantly advanced flow estimation, among which RAFT~\cite{teed2020raft} establishes a strong baseline by combining dense all-pairs correlation with recurrent iterative refinement. Building on this framework, SEA-RAFT~\cite{wang2024sea} improves efficiency and robustness through a simplified update mechanism and a mixed Laplace loss. Recently, FlowSeek~\cite{poggi2025flowseek} further enhances flow estimation by incorporating stronger priors and more efficient architectures, achieving impressive performance on high-quality inputs.

\subsubsection{Geometric correspondence.}
Establishing reliable geometric correspondence is fundamental to many vision tasks. Classical pipelines rely on handcrafted local descriptors~\cite{lowe2004distinctive,bay2006surf}, while learned CNN and transformer models have substantially improved matching robustness~\cite{tian2017l2,mishchuk2017hardnet,detone2018superpoint,sun2021loftr}. However, accurately modeling \emph{dense} correspondences for fine-grained geometric alignment remains challenging, especially under large appearance variations. Recent studies show that diffusion models provide spatially informative representations for correspondence. In particular, DIFT demonstrates that correspondence can emerge from image diffusion features without explicit supervision or task-specific fine-tuning~\cite{tang2023emergent}. Complementary to diffusion features, DINOv2 offers strong semantic representations, and a simple fusion of diffusion and DINOv2 features yields more robust dense correspondences~\cite{oquab2023dinov2,zhang2023tale}. For videos, DiffTrack further reveals that query-key similarities in selected layers of video diffusion transformers encode temporal correspondences across frames~\cite{nam2025emergent}.

\subsubsection{Restoration diffusion model.}

Diffusion models have emerged as a powerful paradigm for image restoration, owing to their rich generative priors, stable optimization, and strong generalization through iterative denoising~\cite{saharia2022image, lin2024diffbir, wu2024seesr, ai2024dreamclear, duan2025dit4sr}. By conditioning on degraded observations, these models recover perceptually sharp and realistic details that GAN-based methods~\cite{ledig2017photo} often fail to capture. However, naively extending image restoration diffusion models to video by processing frames independently leads to temporal flickering and visual inconsistency, as they lack a sufficient cross-frame modeling mechanism~\cite{yang2024motion, zhou2024upscale}. More recently, video diffusion methods~\cite{xie2025star, chen2025doveefficientonestepdiffusion, zhuang2025flashvsr} have been introduced to leverage strong generative priors for temporal modeling. While effective in restoring spatio-temporal content, these approaches incur substantial computational overhead and expose a trade-off between spatial fidelity and temporal coherence~\cite{ho2022video}.
\section{Preliminaries}
\subsection{Optical Flow Estimation}
\label{subsec:flow_pipline}
Modern optical flow methods generally follow a three-stage pipeline. Given two consecutive frames, a \textbf{feature encoder} $\mathcal{E}$ first encodes each frame into a dense feature representation. A \textbf{correlation operator} $\mathcal{C}$ then constructs a cost volume from pairwise similarities between the two feature maps. Finally, an \textbf{iterative update operator} $\mathcal{U}$ refines an initial flow estimate by repeatedly querying the cost volume through a recurrent unit, conditioned on context features that provide per-pixel information about the reference frame. Denoting the overall model as $\mathcal{M}$, this pipeline can be written compactly as:
\begin{equation}
  \mathcal{M} = \mathcal{U} \circ \mathcal{C} \circ \mathcal{E},
  \label{eq:pipeline}
\end{equation}
where $\circ$ denotes function composition. While this pipeline achieves strong accuracy on clean inputs, its performance degrades substantially on low-quality (LQ) videos, where noise, compression, and blur corrupt the extracted features and distort the resulting correlation signal.

\subsection{DiT-based Image Restoration}
Given a paired low-quality and high-quality frame $(\mathbf{I}^{k}_{\mathrm{LQ}},\, \mathbf{I}^{k}_{\mathrm{HQ}})$, both are first encoded into the latent space via a pretrained variational autoencoder (VAE)~\cite{kingma2013auto}:
\begin{equation}
    \mathbf{z}^{k}_{\mathrm{LQ}} = \mathrm{Enc}(\mathbf{I}^{k}_{\mathrm{LQ}}),  \qquad
    \mathbf{z}^{k}_{\mathrm{HQ}} = \mathrm{Enc}(\mathbf{I}^{k}_{\mathrm{HQ}}).
\end{equation}
The diffusion process operates exclusively on the clean latent $\mathbf{z}^{k}_{\mathrm{HQ}}$, while the degraded latent $\mathbf{z}^{k}_{\mathrm{LQ}}$ serves solely as a conditioning signal. Models such as DiT4SR~\cite{duan2025dit4sr} also accept a text prompt as an additional condition; we omit it from our notation for brevity. During training, a noisy latent $\mathbf{z}^{k}_{t}$ is constructed by linearly interpolating between Gaussian noise and the clean target according to a continuous noise level $t \in [0,1]$:
\begin{equation}
    \mathbf{z}^{k}_{t} = (1 - t)\,\boldsymbol{\epsilon} + t\,\mathbf{z}^{k}_{\mathrm{HQ}},
    \qquad \boldsymbol{\epsilon} \sim \mathcal{N}(\mathbf{0},\, \mathbf{I}).
\end{equation}
The DiT-based denoising network $\mathcal{D}$ is then trained to predict the velocity field along this interpolation path, conditioned on the degraded latent:
\begin{equation}
    \mathbf{v}^{k}_{t} = \mathcal{D}(\mathbf{z}^{k}_{t},\, t \mid \mathbf{z}^{k}_{\mathrm{LQ}}).
\end{equation}
Under the rectified flow formulation~\cite{liu2022flow}, the ground-truth velocity is obtained by differentiating $\mathbf{z}^{k}_{t}$ with respect to $t$:
\begin{equation}
    \frac{d\mathbf{z}^{k}_{t}}{dt}
    = \frac{d}{dt}\!\big((1 - t)\,\boldsymbol{\epsilon} + t\,\mathbf{z}^{k}_{\mathrm{HQ}}\big)
    = \mathbf{z}^{k}_{\mathrm{HQ}} - \boldsymbol{\epsilon}.
\end{equation}
The model is thus trained by minimizing the flow-matching objective:
\begin{equation}
    \mathcal{L}_{\mathrm{diff}} =
    \mathbb{E}_{k,\,t,\,\boldsymbol{\epsilon}}
    \!\left[
    \left\|
    \mathbf{v}^{k}_{t} - (\mathbf{z}^{k}_{\mathrm{HQ}} - \boldsymbol{\epsilon})
    \right\|_2^2
    \right].
    \label{eq:diff_loss}
\end{equation}
At inference, the model starts from pure noise and iteratively denoises the latent using the learned velocity field; the final restored image is then obtained by decoding the result with the VAE decoder.


\begin{figure}[t]
  \centering
  \includegraphics[width=1\linewidth]{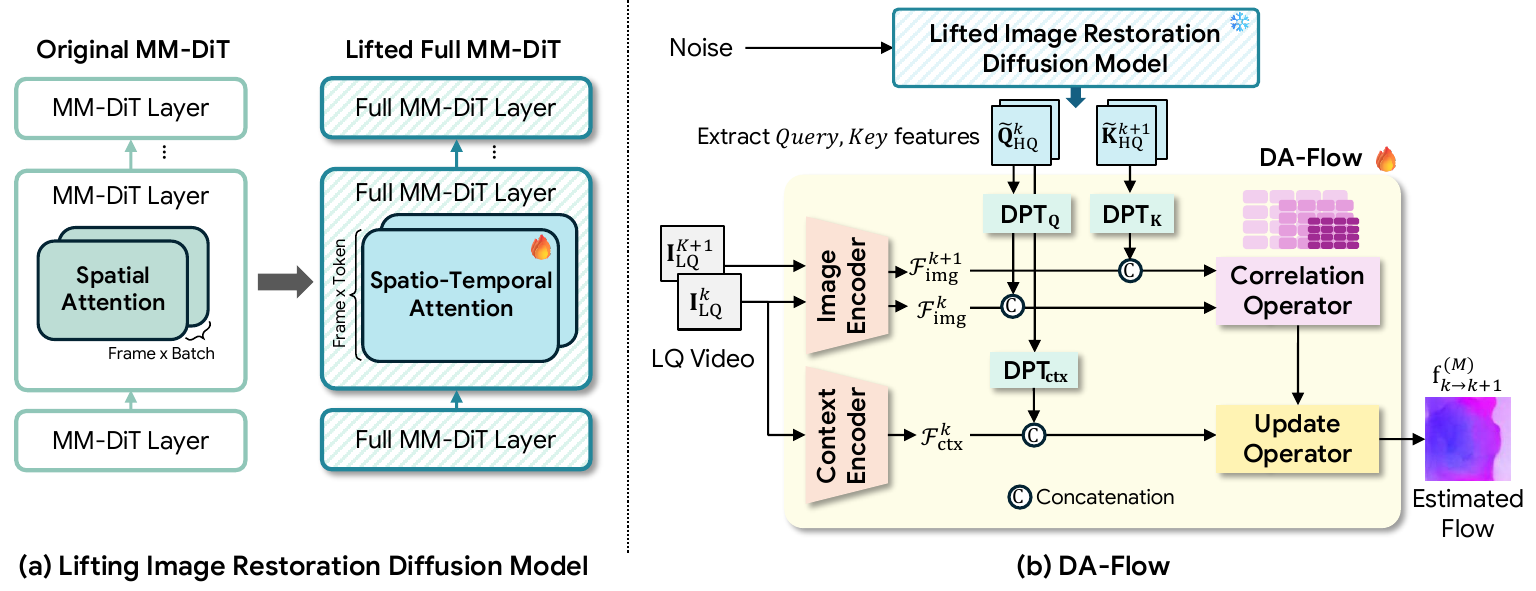}
  \caption{\textbf{Overall architecture of \ourmodel.}}
  \label{figs:main_figure}
  \vspace{-10pt}
\end{figure}

\section{Method}
\label{sec:method}
In this section, we present our approach to degradation-aware optical flow estimation. We begin by describing how a pretrained DiT-based image restoration model is lifted to the video domain through full spatio-temporal attention in Sec.~\ref{subsec:lifting}. We then analyze the geometric correspondence encoded in the diffusion features across different layers, identifying which layers yield the most correspondence-ready representations in Sec.~\ref{subsec:analysis}. Based on these findings, we introduce \textbf{\ourmodel}, our degradation-aware optical flow model that leverages the selected diffusion features to estimate reliable motion from corrupted inputs in Sec.~\ref{subsec:flow_head}.

\subsection{Problem Formulation}
We introduce a new task of \textbf{Degradation-Aware Optical Flow}, which aims to estimate accurate flow from corrupted videos. Let $\mathbf{V}_{\text{LQ}}$ and $\mathbf{V}_{\text{HQ}}$ denote a low-quality (LQ) video and its corresponding high-quality (HQ) video, respectively, each represented as a sequence of $N$ RGB frames $\{\mathbf{I}^{i}\}_{i=1}^{N}$ with $\mathbf{I}^{i} \in \mathbb{R}^{3 \times H \times W}$. Our goal is to learn a degradation-aware optical flow model $\mathcal{M}$ that reliably estimates motion from LQ inputs. For a pair of consecutive frames indexed by $k$ and $k\!+\!1$, where $k \in \{1, \dots, N\!-\!1\}$, the model estimates:
\begin{equation}
  \widehat{\mathbf{f}}_{k \to k+1}
  = \mathcal{M}\!\bigl(\mathbf{I}_{\text{LQ}}^{k},\;\mathbf{I}_{\text{LQ}}^{k+1}\bigr)
  \approx \mathbf{f}_{k \to k+1}^{*},
  \label{eq:flow_estimation}
\end{equation}
where $\mathbf{f}_{k \to k+1}^{*}$ denotes the ground-truth flow between frames $k$ and $k\!+\!1$. Among the three stages in Eq.~\ref{eq:pipeline}, the feature encoder $\mathcal{E}$ is most directly affected by input degradation, as corrupted pixels lead to unreliable features that propagate errors into all downstream stages. We therefore focus on building a degradation-aware feature encoder that produces robust, correspondence-ready representations from LQ inputs, while keeping $\mathcal{C}$ and $\mathcal{U}$ unchanged from existing architectures.

\subsection{Lifting Image Restoration Model}
\label{subsec:lifting}
The DiT-based image restoration model operates independently on each frame, providing strong per-frame restoration capability but lacking any mechanism for temporal modeling. To preserve this strong generative prior while enabling temporal reasoning, we extend the model with full spatio-temporal attention over tokens across multiple frames.

\subsubsection{Multi-modal attention in MM-DiT.}
Our backbone is based on the Multi-Modal Diffusion Transformer (MM-DiT)~\cite{esser2024scaling}. A straightforward way to apply this image-level model to video is to fold the temporal dimension into the batch axis. Each of the F frames in a batch of B video clips is then processed independently, yielding BF separate sequences of T patchified tokens with channel dimension $C$. Under this scheme, MM-DiT processes three modality-specific token sequences per frame through a Multi-Modal Attention (MM-Attention) mechanism:
(i) $\mathbf{F}_{\text{HQ}} \in \mathbb{R}^{(BF) \times T \times C}$, latent tokens representing the current denoising state;
(ii) $\mathbf{F}_{\text{LQ}} \in \mathbb{R}^{(BF) \times T \times C}$, conditioning tokens from the degraded input; and
(iii) $\mathbf{F}_{\text{Text}} \in \mathbb{R}^{(BF) \times T \times C}$, text tokens encoding semantic priors.
Within each block, modality-specific projections produce queries, keys, and values:
\begin{equation}
\begin{gathered}
    (\mathbf{Q}_{m}, \mathbf{K}_{m}, \mathbf{V}_{m})
    = (\mathbf{F}_{m}\mathbf{W}_{Q}^{m},\;
       \mathbf{F}_{m}\mathbf{W}_{K}^{m},\;
       \mathbf{F}_{m}\mathbf{W}_{V}^{m}),
     \quad m \in \{\text{HQ}, \text{LQ}, \text{Text}\},
\end{gathered}
\end{equation}
which are concatenated along the token dimension to form $\mathbf{Q}, \mathbf{K}, \mathbf{V} \in \mathbb{R}^{(BF) \times 3T \times C}$ for joint attention. However, since the temporal dimension remains folded into the batch axis, MM-Attention is applied independently per frame, preventing the model from capturing inter-frame correspondences.

\subsubsection{Full spatio-temporal attention.}
To enable inter-frame reasoning for our task, we reshape each modality stream from $\mathbf{F}_{m} \in \mathbb{R}^{(BF) \times T \times C}$ to $\tilde{\mathbf{F}}_{m} \in \mathbb{R}^{B \times (FT) \times C}$, concatenating all spatial tokens across frames into a single sequence per video. Modality-specific projections and concatenation then yield spatio-temporal queries, keys, and values
\begin{equation}
\tilde{\mathbf{Q}}, \tilde{\mathbf{K}}, \tilde{\mathbf{V}}
\in \mathbb{R}^{B \times (3FT) \times C},
\end{equation}
and full spatio-temporal MM-Attention is computed as
\begin{equation}
\mathrm{MM\text{-}Attn}
= \mathrm{softmax}\!\left(
\frac{\tilde{\mathbf{Q}}\tilde{\mathbf{K}}^{\top}}{\sqrt{C}}
\right)\tilde{\mathbf{V}},
\end{equation}
where each token now attends to all spatial tokens across all frames and modalities, enabling explicit inter-frame reasoning. With this full spatio-temporal MM-Attention applied to all layers, we finetune the lifted diffusion model $\mathcal{D}_{\phi}$ on the YouHQ training dataset~\cite{zhou2024upscale}. After training, we use this lifted model as the feature encoder $\mathcal{E}$ in \ourmodel, leveraging its temporally-aware diffusion features for flow estimation. Further details are provided in Sec.~\ref{subsec:exp_setup} and Appendix~\ref{appendix:imple_details}.

\subsection{Diffusion Feature Analysis}
\label{subsec:analysis}
A remaining question is which intermediate representations to extract from the lifted model for flow estimation. Recent work such as DiffTrack~\cite{nam2025emergent} shows that query and key features from full spatio-temporal attention layers in video diffusion models exhibit strong geometric correspondence. Motivated by this finding, we extract attention features from the full spatio-temporal MM-Attention layers introduced during lifting. Specifically, given a consecutive frame pair $(k, k\!+\!1)$, we take the query feature from frame $k$ and the key feature from frame $k\!+\!1$ in the HQ diffusion branch:
\begin{equation}
\begin{aligned}
\tilde{\mathbf{Q}}_{\text{HQ}}^{k},\
\tilde{\mathbf{K}}_{\text{HQ}}^{k+1}
\in \mathbb{R}^{B \times T \times C}.
\end{aligned}
\label{eq:diff_qk_def}
\end{equation}
Note that, unlike prior works~\cite{tang2023emergent, nam2025emergent} that inject input images into the generation branch at a specific noise level $t$, our features are extracted during the iterative denoising process, and we accordingly analyze them across denoising timesteps rather than at a single predetermined noise level. A further comparison with an alternative feature type is provided in Appendix~\ref{appendix:feature_analysis}.
\begin{figure}[t]
    \centering
        \begin{subfigure}[t]{0.48\linewidth}
            \centering
            \includegraphics[width=\linewidth]{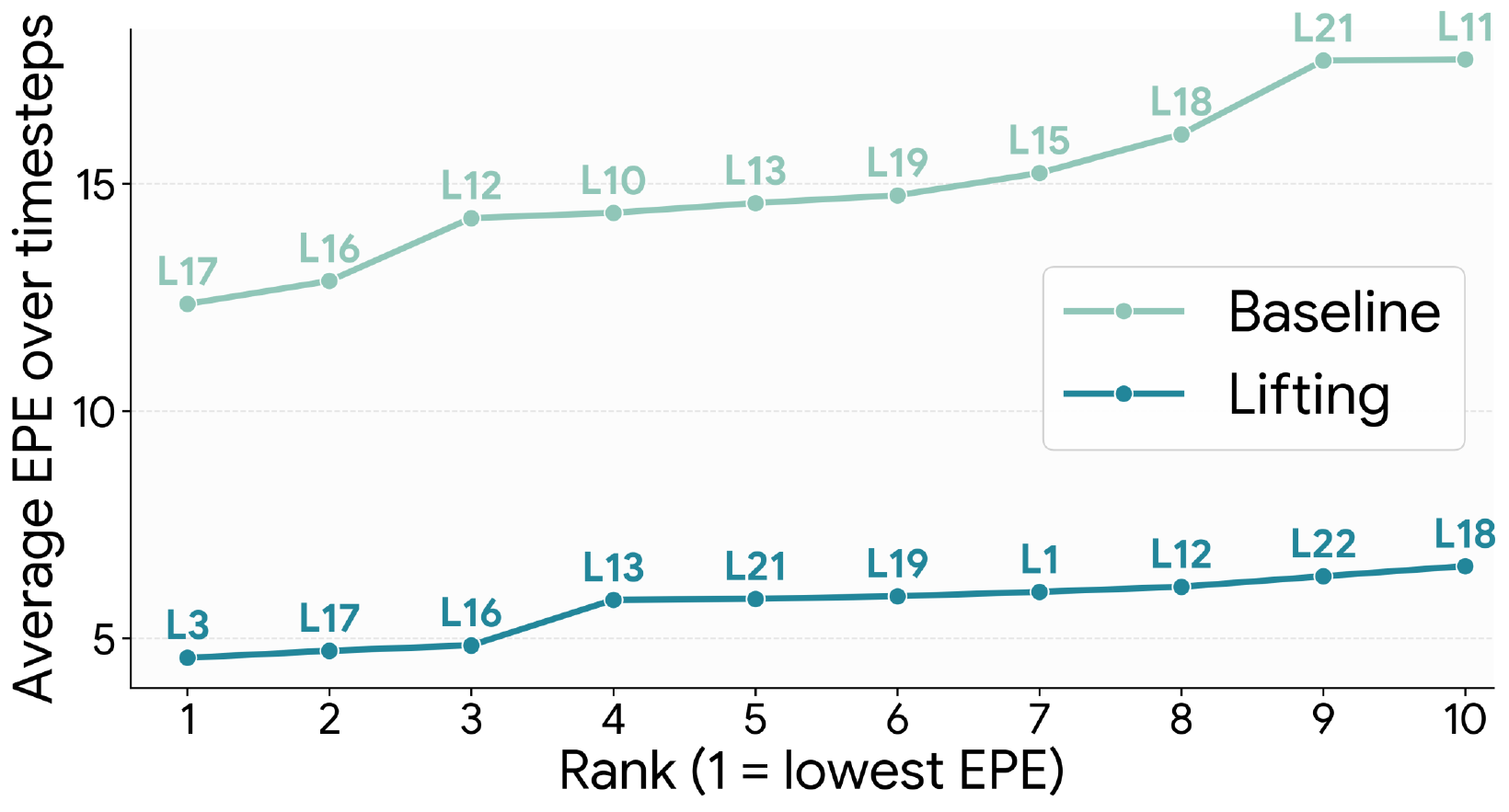}
            \caption{Top-10 Layer-wise Average EPE}
            \label{fig:sub1}
    \end{subfigure}
    \hfill
    \begin{subfigure}[t]{0.48\linewidth}
        \centering
        \includegraphics[width=\linewidth]{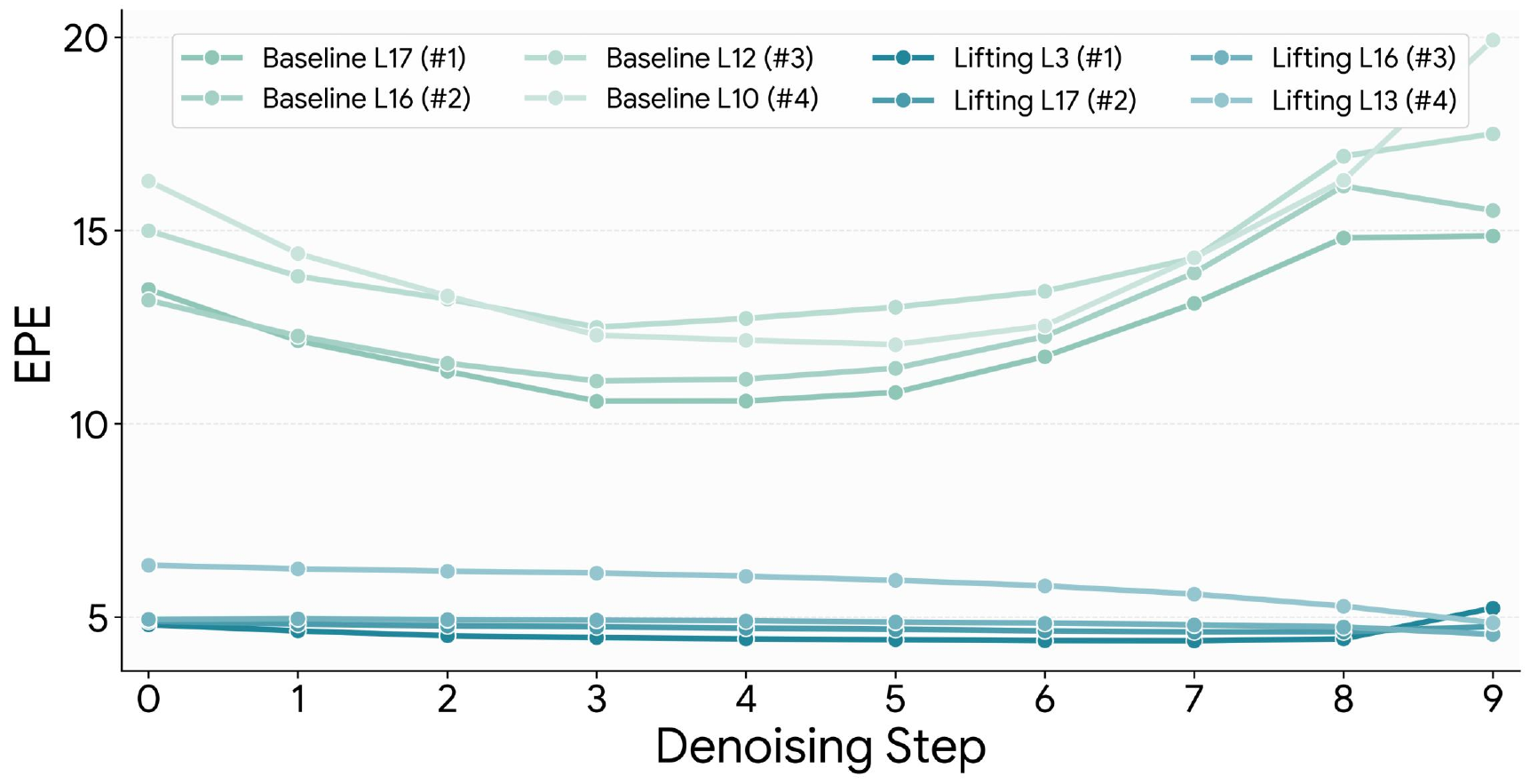}
        \caption{Top-4 Layer EPE over Denoising Steps}
        \label{fig:sub2}
    \end{subfigure}
    
    \caption{\textbf{Comparison of zero-shot geometric correspondence between Baseline and Lifting features.} (a) Top-10 layers ranked by timestep-averaged EPE (lower is better). Lifting consistently achieves lower EPE across all ranks. (b) EPE over denoising steps for the top-4 layers of each method. Baseline features show high sensitivity to the denoising step, while Lifting features remain stable across the denoising steps.}
    \label{fig:main}\vspace{-10pt}
\end{figure}
\subsubsection{Evaluation protocol.}
To assess the zero-shot geometric correspondence of these diffusion features, we evaluate them through direct flow estimation without any task-specific training. Each feature has $T = h \times w$ tokens corresponding to the spatial dimensions of the latent space. For a single frame pair $(k, k\!+\!1)$, the extracted features $\tilde{\mathbf{Q}}_{\text{HQ}}^{k},\, \tilde{\mathbf{K}}_{\text{HQ}}^{k+1} \in \mathbb{R}^{T \times C}$ are reshaped to $\mathbb{R}^{h \times w \times C}$, from which we construct a cost volume $\mathbf{C} \in \mathbb{R}^{h \times w \times h \times w}$ by computing pairwise dot-product similarity:
\begin{equation}
    \mathbf{C}(i, j) =
    \tilde{\mathbf{Q}}_{\text{HQ}}^{k}(i) \cdot \tilde{\mathbf{K}}_{\text{HQ}}^{k+1}(j).
    \label{eq:cost_volume}
\end{equation}
A flow field $\widehat{\mathbf{f}}_{k \to k+1} \in \mathbb{R}^{h \times w \times 2}$ is then obtained via $\widehat{\mathbf{f}}_{k \to k+1} = \mathrm{softargmax}(\mathbf{C})$ and upsampled to the original image resolution $H \times W$. To evaluate this zero-shot prediction, we obtain a pseudo ground-truth flow $\mathbf{f}^{*}_{k \to k+1}$ by applying a pretrained optical flow model to the corresponding HQ frame pair $(\mathbf{I}_{\text{HQ}}^{k},\, \mathbf{I}_{\text{HQ}}^{k+1})$, which serves as the reference for measuring correspondence accuracy. We report End-Point Error (EPE) on LQ--HQ video pairs from the YouHQ40~\cite{zhou2024upscale} validation set.

\subsubsection{Results.}
We compare two configurations: the \textit{Baseline}, which applies full spatio-temporal attention but is not finetuned, and \textit{Lifting}, which is finetuned on YouHQ as described in Sec.~\ref{subsec:lifting}. As shown in Fig.~\ref{fig:sub1}, the lifted model achieves consistently lower EPE than the baseline across all layer ranks, confirming that finetuning with full spatio-temporal attention enables the model to learn inter-frame correspondences absent in the untrained baseline. The lifted features also remain stable across the entire denoising trajectory in Fig.~\ref{fig:sub2}, in contrast to the baseline which exhibits high sensitivity to the extraction timestep. These results demonstrate that the lifted features possess superior geometric correspondence quality, and provide the basis for selecting which layers to extract features for \ourmodel, as detailed in Sec.~\ref{subsec:flow_head}. More detailed analyses are provided in Appendix~\ref{appendix:feature_analysis}.

\subsection{\ourmodel}
\label{subsec:flow_head}
Building upon the lifting architecture in Sec.~\ref{subsec:lifting} and the empirical analysis in Sec.~\ref{subsec:analysis}, we introduce \ourmodel, a degradation-aware optical flow model built on top of RAFT~\cite{teed2020raft}. As illustrated in Fig.~\ref{figs:main_figure}, \ourmodel\ retains the original correlation operator $\mathcal{C}$ and iterative update operator $\mathcal{U}$, while incorporating the lifted diffusion model $\mathcal{D_\phi}$ alongside a conventional feature encoder $\mathcal{E}$. The overall pipeline can be written as:
\begin{equation}
  \mathcal{M}_{\theta} = \mathcal{U} \circ \mathcal{C} \circ (\texttt{Up}(\mathcal{D_\phi}),\; \mathcal{E}),
  \label{eq:our_pipeline}
\end{equation}
where $\texttt{Up}$ denotes a learnable upsampling stage that maps the coarse diffusion features to a resolution compatible with $\mathcal{E}$. In the following, we describe each component in detail.

\subsubsection{Feature upsampling.} 
The diffusion features produced by $\mathcal{D}_{\phi}$ lie on a coarse spatial grid at $1/16$ of the input resolution. Directly passing them to the correlation operator $\mathcal{C}$ limits the quality of the resulting 4D cost volume, since accurate flow estimation requires fine-grained spatial details for precise boundary localization. We now describe the upsampling stage $\texttt{Up}$ in Eq.~\ref{eq:our_pipeline} that addresses this resolution gap.

Specifically, we aggregate diffusion features from the top-$L$ layers that exhibit the strongest geometric correspondence quality, as identified in Sec.~\ref{subsec:analysis}. The aggregated features are then passed through DPT-based upsampling heads~\cite{ranftl2021vision} to recover higher-resolution feature maps. Since the query and key features from the diffusion attention already encode distinct representations, we employ separate DPT heads to preserve this distinction: a query head and a key head produce correspondence features for cost volume construction, while a context head generates spatial conditioning features for the iterative update operator $\mathcal{U}$. Formally, given frame index $k$, the upsampled features are obtained as:
\begin{equation}
\begin{aligned}
\mathcal{F}_{\text{Q}}^{k,\uparrow}
&= \text{DPT}_\text{Q}\!\left(\{\tilde{\mathbf{Q}}_{\text{HQ}}^{k,l}\}_{l=1}^{L}\right), \\
\mathcal{F}_{\text{K}}^{k+1,\uparrow}
&= \text{DPT}_\text{K}\!\left(\{\tilde{\mathbf{K}}_{\text{HQ}}^{k+1,l}\}_{l=1}^{L}\right), \\
\mathcal{F}_{\text{ctx}}^{k,\uparrow}
&= \text{DPT}_{\text{ctx}}\!\left(\{\tilde{\mathbf{Q}}_{\text{HQ}}^{k,l}\}_{l=1}^{L}\right),
\end{aligned}
\label{eq:dpt_upsample}
\end{equation}
where $l$ indexes the selected diffusion layers. All upsampled features share a common spatial resolution of $H/8 \times W/8$, with query and key features having channel dimension $c$ and context features having channel dimension $c'$.

\begin{table}[t]
    \centering
    \caption{\textbf{Quantitative comparison on Sintel, Spring, and TartanAir.} All methods are evaluated using End-Point Error (EPE) and outlier rates at 1px, 3px, and 5px thresholds. \colorbox{best}{Best} and \colorbox{secondbest}{second best} results are highlighted.}
    \resizebox{\textwidth}{!}{%
        \begin{tabular}{l|cccc|cccc|cccc}
        \toprule
        \multirow{2}{*}{\textbf{Model}} &
        \multicolumn{4}{c|}{\textbf{Sintel}} &
        \multicolumn{4}{c|}{\textbf{Spring}} &
        \multicolumn{4}{c}{\textbf{TartanAir}} \\
        \cmidrule(lr){2-5}\cmidrule(lr){6-9}\cmidrule(lr){10-13}
        & {EPE}$\downarrow$ & {1px}$\downarrow$ & {3px}$\downarrow$ & {5px}$\downarrow$
        & {EPE}$\downarrow$ & {1px}$\downarrow$ & {3px}$\downarrow$ & {5px}$\downarrow$
        & {EPE}$\downarrow$ & {1px}$\downarrow$ & {3px}$\downarrow$ & {5px}$\downarrow$ \\
        \midrule
        RAFT~\cite{teed2020raft}
            & 10.693 & 62.91 & 37.24 & 28.63
            & 3.944 & \cellcolor{secondbest}39.82 & \cellcolor{secondbest}18.65 & \cellcolor{secondbest}11.98
            & 9.487 & \cellcolor{secondbest}75.17 & \cellcolor{secondbest}42.96 & \cellcolor{secondbest}30.04 \\
        SEA-RAFT~\cite{wang2024sea}
            & \cellcolor{secondbest}10.185 & \cellcolor{secondbest}59.56 & \cellcolor{secondbest}34.46 & \cellcolor{secondbest}26.15
            & \cellcolor{secondbest}2.703 & 41.51 & 19.31 & 12.11
            & \cellcolor{secondbest}8.316 & 77.85 & 45.76 & 32.15 \\
        FlowSeek~\cite{poggi2025flowseek}
            & 10.241 & 64.08 & 40.71 & 31.83
            & 2.861 & 41.53 & 19.16 & 12.18
            & \cellcolor{best}7.694 & 76.96 & 45.20 & 32.00 \\
        \ourmodel
            & \cellcolor{best}6.912 & \cellcolor{best}55.80 & \cellcolor{best}28.10 & \cellcolor{best}20.91
            & \cellcolor{best}2.207 & \cellcolor{best}30.95 & \cellcolor{best}13.87 & \cellcolor{best}8.91
            & 8.866 & \cellcolor{best}72.35 & \cellcolor{best}37.61 & \cellcolor{best}25.40 \\
        \bottomrule
        \end{tabular}
    }
    \label{tab:main_quan}\vspace{-10pt}
\end{table}
\subsubsection{Hybrid feature encoding.}
While the upsampled diffusion features $\mathcal{F}_{\text{Q}}^{k,\uparrow}$, $\mathcal{F}_{\text{K}}^{k+1,\uparrow}$, and $\mathcal{F}_{\text{ctx}}^{k,\uparrow}$ provide strong degradation-aware representations, they lack fine-grained spatial localization due to their globally aggregated nature. To compensate, we incorporate the conventional feature encoder $\mathcal{E}$ from Eq.~\ref{eq:our_pipeline}, which preserves local spatial details through its convolutional architecture. Following RAFT, $\mathcal{E}$ consists of an image encoder $\mathcal{E}_{\text{img}}$ applied to all input frames and a context encoder $\mathcal{E}_{\text{ctx}}$ applied only to the reference frame:
\begin{equation}
\begin{aligned}
\mathcal{F}_{\text{img}}^{k},\; \mathcal{F}_{\text{img}}^{k+1}
    &= \mathcal{E}_{\text{img}}(\mathbf{I}_{\text{LQ}}^{k}),\; \mathcal{E}_{\text{img}}(\mathbf{I}_{\text{LQ}}^{k+1}), \\
\mathcal{F}_{\text{ctx}}^{k} 
    &= \mathcal{E}_{\text{ctx}}(\mathbf{I}_{\text{LQ}}^{k}),
\end{aligned}
\label{eq:raft_enc}
\end{equation}
where all encoder features have spatial resolution $H/8 \times W/8$. The diffusion and CNN features are then concatenated along the channel dimension to form the hybrid representations used for cost volume construction and iterative updates.
The hybrid feature maps are then formed by concatenating the diffusion and CNN features along the channel dimension:
\begin{equation}
    \begin{aligned}
        \mathcal{F}^{k} &= \text{Concat}(\mathcal{F}_{\text{img}}^{k},\; \mathcal{F}_{\text{Q}}^{k,\uparrow}), \\
        \mathcal{F}^{k+1} &= \text{Concat}(\mathcal{F}_{\text{img}}^{k+1},\; \mathcal{F}_{\text{K}}^{k+1,\uparrow}), \\
        \mathcal{F}_{\text{h-ctx}}^{k} &= \text{Concat}(\mathcal{F}_{\text{ctx}}^{k},\; \mathcal{F}_{\text{ctx}}^{k,\uparrow}),
    \end{aligned}
    \label{eq:concat}
\end{equation}
where $\mathcal{F}^{k}$ and $\mathcal{F}^{k+1}$ are used to construct the correlation volume via $\mathcal{C}$, and $\mathcal{F}_{\text{h-ctx}}^{k}$ provides spatial conditioning for the iterative update operator $\mathcal{U}$. The correlation volume and context features are then processed through $\mathcal{U}$, which produces a sequence of refined flow estimates:
\begin{equation}
    \{\mathbf{f}_{k \rightarrow k+1}^{(i)}\}_{i=0}^{M},
    \label{eq:pred_flow}
\end{equation}
where $\mathbf{f}^{(i)} \in \mathbb{R}^{H \times W \times 2}$ and $M$ denotes the number of recurrent refinement steps. The final flow field $\mathbf{f}_{k \rightarrow k+1}^{(M)}$ is taken from the last iteration.

\subsubsection{Loss function.}
\label{subsec:training}
Since obtaining ground-truth optical flow for real-world degraded videos is infeasible, we generate pseudo ground-truth labels from the same YouHQ dataset~\cite{zhou2024upscale} used to train the lifted diffusion model in Sec.~\ref{subsec:lifting}. Following the protocol in Sec.~\ref{subsec:analysis}, we apply a pretrained optical flow model to the HQ frame pairs to obtain pseudo ground-truth flow $\mathbf{f}^{*}_{k \rightarrow k+1}$, while the corresponding LQ frames serve as input to \ourmodel. We optimize $\mathcal{M}_{\theta}$ using the standard multi-scale flow loss:
\begin{equation}
    \mathcal{L}_\text{flow} = \sum_{i=1}^{M} \gamma^{M-i} \left\| \mathbf{f}_{k \rightarrow k+1}^{(i)} - \mathbf{f}^{*}_{k \rightarrow k+1} \right\|_1,
    \label{eq:flow_loss}
\end{equation}
where $\gamma$ is a weight decay factor and $M$ denotes the number of refinement iterations.

\section{Experiment}
\label{sec:exp}

\begin{figure}[t]
  \centering
  \includegraphics[width=1\linewidth]{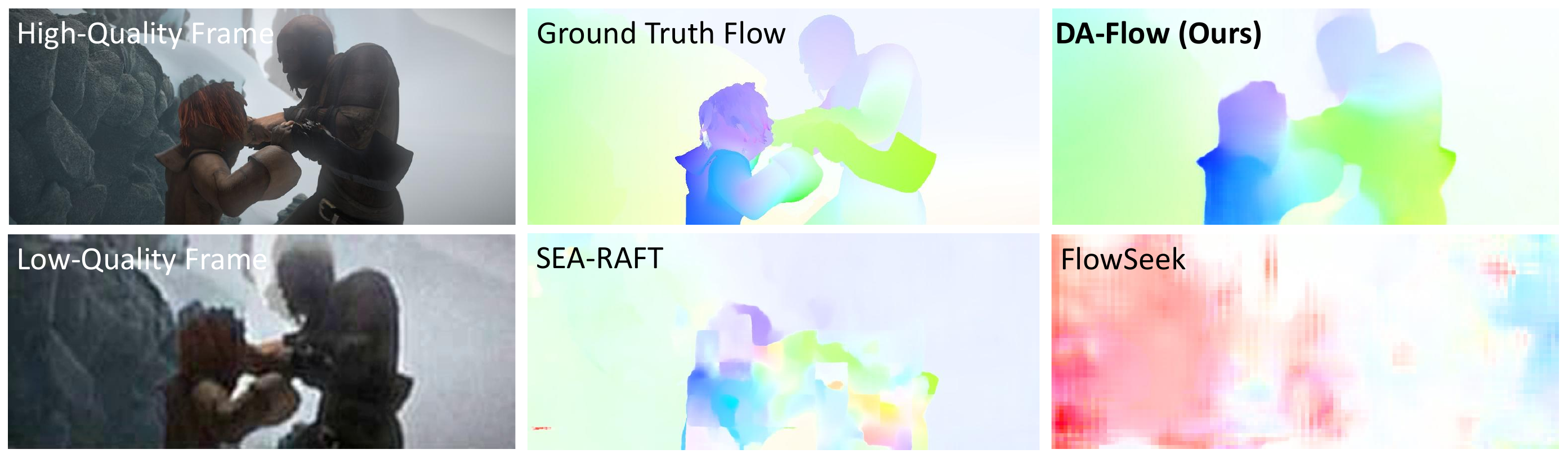}
  \caption{\textbf{Qualitative results on Sintel~\cite{butler2012naturalistic}.}}
  \label{figs:qual_sintel}
  \vspace{-10pt}
\end{figure}
\begin{figure}[t]
  \centering
  \includegraphics[width=1\linewidth]{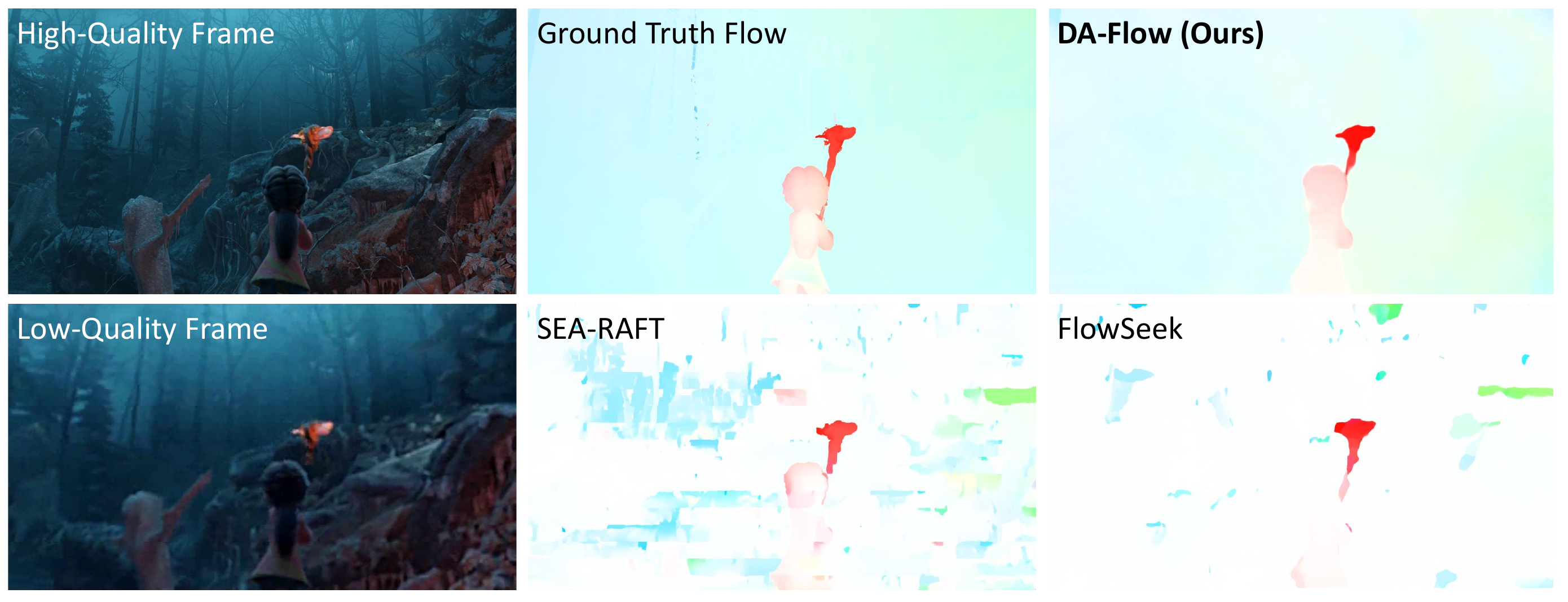}
  \caption{\textbf{Qualitative results on Spring~\cite{Mehl2023_Spring}.}}
  \label{figs:qual_spring}
  \vspace{-10pt}
\end{figure}
\subsection{Experimental Setup}
\label{subsec:exp_setup}
\subsubsection{Implementation details.}

We train our model in two stages. In the first stage, the lifted diffusion model $\mathcal{D}_{\phi}$ is trained with the diffusion loss $\mathcal{L}_{\text{diff}}$ on clips of $F=3$ consecutive frames. In the second stage, $\mathcal{D}_{\phi}$ is frozen, and the flow estimation pipeline $\mathcal{M}_{\theta}$ is trained with the flow loss $\mathcal{L}_{\text{flow}}$ using $M=12$ refinement iterations. For correspondence estimation, we extract diffusion features from four full spatio-temporal attention layers $\{3, 13, 16, 17\}$ ($L=4$), selected based on the analysis in Sec.~\ref{subsec:analysis}. Both stages are trained for 20K steps each with a batch size of 32 on 4 NVIDIA H100 GPUs and a learning rate of $5\times10^{-5}$.

\label{subsec:train_dataset}
\subsubsection{Training dataset.}
We train \ourmodel\ on YouHQ~\cite{zhou2024upscale}, a high-resolution video dataset comprising 38,576 diverse videos sourced from YouTube. With an average resolution of 1080$\times$1920 and 32 frames per clip, the dataset spans various scenarios including street views, human portraits, animals, and other categories. To generate corresponding LQ pairs, we follow the degradation pipeline of RealBasicVSR~\cite{chan2022investigating}, which applies the Real-ESRGAN~\cite{wang2021real} degradation model at the frame level, followed by video compression over the entire sequence. Since ground-truth optical flow is not available, we generate pseudo ground-truth flow from the HQ frame pairs using SEA-RAFT~\cite{wang2024sea}. The same degradation and pseudo ground-truth generation setup is applied to the YouHQ40~\cite{zhou2024upscale} validation split used for the feature analysis in Sec.~\ref{subsec:analysis}. During training, input frames are randomly cropped to $512\times512$.

\begin{figure}[t]
  \centering
  \includegraphics[width=1\linewidth]{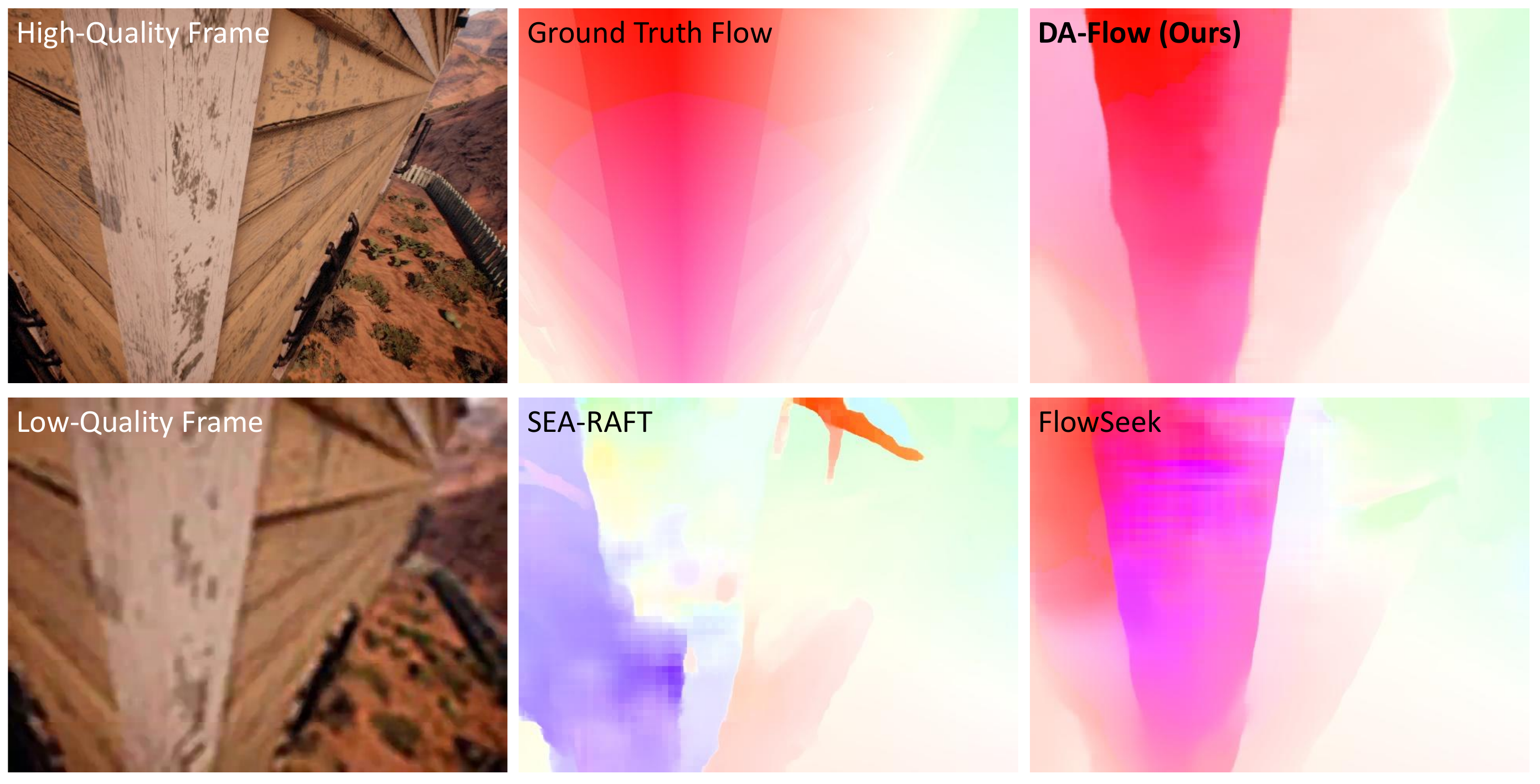}
  \caption{\textbf{Qualitative results on TartanAir~\cite{wang2020tartanair}.}}
  \label{figs:qual_tartanair}
  \vspace{-10pt}
\end{figure}
\subsubsection{Evaluation datasets and metrics.}
We evaluate \ourmodel\ on three optical flow benchmarks. Spring~\cite{Mehl2023_Spring} is a large-scale synthetic dataset providing dense ground-truth flow with highly detailed scenes and complex motion patterns. The Sintel~\cite{butler2012naturalistic} train set offers two rendering passes (clean and final) with varying levels of motion blur and atmospheric effects. Following FlowSeek~\cite{poggi2025flowseek}, we additionally construct an evaluation set from the TartanAir~\cite{wang2020tartanair} validation split using the same setup. For all benchmarks, LQ inputs are generated using the same degradation pipeline employed during training to ensure consistency. We report End-Point Error (EPE) and the percentage of outlier pixels with errors exceeding 1px, 3px, and 5px thresholds as evaluation metrics. The number of iterative updates for all baseline models is set to 12, ensuring a fair comparison. \ourmodel\ performs 10 denoising steps during inference.

\subsection{Quantitative Results}
Tab.~\ref{tab:main_quan} compares \ourmodel\ against existing optical flow methods on three benchmarks under synthetic degradations. On Sintel and Spring, \ourmodel\ achieves the best performance across all metrics, reducing EPE by a clear margin over the strongest baseline. This highlights the effectiveness of leveraging degradation-aware diffusion features for flow estimation under corrupted inputs. On TartanAir, \ourmodel\ achieves the best outlier rates at all thresholds (1px, 3px, 5px), while showing a higher EPE than FlowSeek. This discrepancy can be attributed to a small number of pixels with large displacement errors; these outlier pixels disproportionately inflate the average endpoint error. The consistently lower outlier rates suggest that \ourmodel\ produces more accurate estimates over the majority of pixels. Additional results can be found in Appendix~\ref{appendix:additional_results}.

\subsection{Qualitative Results}
\label{sec:qualitative}

We present qualitative comparisons against SEA-RAFT and FlowSeek on Sintel, Spring, and TartanAir in Fig.~\ref{figs:qual_sintel}, Fig.~\ref{figs:qual_spring}, and Fig.~\ref{figs:qual_tartanair}. Across all benchmarks, baseline methods produce noisy and inconsistent flow fields under degraded inputs, with artifacts concentrated around motion boundaries and fine-grained structures, which are an expected failure mode for methods designed for clean inputs. DA-Flow, by contrast, consistently recovers sharp, coherent flow fields that closely match the ground truth, successfully localizing motion boundaries on Sintel, maintaining structural coherence over complex scenes on Spring, and producing cleaner estimates under large displacements on TartanAir. These results demonstrate that the degradation-aware priors encoded in the lifted diffusion features allow DA-Flow to reason about scene geometry even when the underlying visual evidence is severely corrupted. More qualitative results are provided in Appendix~\ref{appendix:quals}.

\begin{table*}[t]
  \centering
  \caption{\textbf{Ablation on feature source across denoising steps.} Baseline* uses the same \ourmodel\ architecture but extracts features from the full-attention model before finetuning. \textbf{Bold} indicates the better result at each step.}
  \label{tab:ablation_stepwise}
  \small
  \setlength{\tabcolsep}{5pt}
  \resizebox{\textwidth}{!}{%
  \begin{tabular}{lllcccccccccc}
    \toprule
     &  &  & \multicolumn{10}{c}{Step} \\
    \cmidrule(lr){4-13}
    Dataset & Metric & Method & 0 & 1 & 2 & 3 & 4 & 5 & 6 & 7 & 8 & 9 \\
    \midrule
    \multirow{4}{*}{Sintel~\cite{butler2012naturalistic}} & \multirow{2}{*}{EPE $\downarrow$} & Baseline* & 7.4145 & 7.4542 & 7.5076 & 7.4703 & 7.4124 & 7.4897 & 7.5243 & 7.5752 & 7.5709 & 7.6883 \\
     &  & \ourmodel & \textbf{7.0210} & \textbf{6.7809} & \textbf{6.7196} & \textbf{6.7160} & \textbf{6.7433} & \textbf{6.7605} & \textbf{6.8029} & \textbf{6.8736} & \textbf{7.0641} & \textbf{7.6397} \\
    \cmidrule(lr){2-13}
     & \multirow{2}{*}{1px $\downarrow$} & Baseline* & 59.47 & 59.26 & 59.50 & 59.26 & 59.60 & 60.16 & 59.85 & 60.80 & 61.02 & 61.79 \\
     &  & \ourmodel & \textbf{57.04} & \textbf{56.81} & \textbf{56.39} & \textbf{55.97} & \textbf{55.52} & \textbf{55.12} & \textbf{54.79} & \textbf{54.72} & \textbf{54.64} & \textbf{57.03} \\
    \midrule
    \multirow{4}{*}{Spring~\cite{Mehl2023_Spring}} & \multirow{2}{*}{EPE $\downarrow$} & Baseline* & 2.3269 & 2.3457 & 2.2535 & 2.3073 & 2.2158 & \textbf{2.2030} & 2.2036 & 2.2148 & 2.2066 & 2.2343 \\
     &  & \ourmodel & \textbf{2.2902} & \textbf{2.2119} & \textbf{2.2069} & \textbf{2.2026} & \textbf{2.2011} & 2.2043 & \textbf{2.2008} & \textbf{2.1928} & \textbf{2.1917} & \textbf{2.1720} \\
    \cmidrule(lr){2-13}
     & \multirow{2}{*}{1px $\downarrow$} & Baseline* & 32.14 & 32.74 & 32.45 & 32.72 & 32.23 & 32.24 & 32.15 & 32.63 & 32.24 & 31.86 \\
     &  & \ourmodel & \textbf{30.49} & \textbf{31.01} & \textbf{31.06} & \textbf{30.99} & \textbf{31.05} & \textbf{31.23} & \textbf{31.23} & \textbf{31.06} & \textbf{30.88} & \textbf{30.54} \\
    \bottomrule
  \end{tabular}%
  }\vspace{-10pt}
\end{table*}

\subsection{Ablation Study}

\subsubsection{Comparison with Baseline.}
To isolate the contribution of the lifted diffusion features, we construct a baseline variant (Baseline*) that uses the same \ourmodel\ architecture but replaces the lifted features with those from the untrained full-attention model. Following the layer-wise analysis in Sec.~\ref{subsec:analysis}, we select the top-4 layers for each model: \{10, 12, 16, 17\} for Baseline* and \{3, 13, 16, 17\} for \ourmodel. Tab.~\ref{tab:ablation_stepwise} reports the results across all denoising steps on Sintel and Spring. On Sintel, \ourmodel\ consistently outperforms Baseline* at every step in both EPE and 1px outlier rate. In Spring, the two methods perform comparably in EPE, while \ourmodel\ maintains a consistent advantage in 1px outlier rate across all steps. In fact, \ourmodel\ achieves lower EPE at nearly every step, except a single step where the two methods are virtually tied. The advantage is more pronounced in 1px outlier rate, where \ourmodel\ consistently outperforms Baseline* across all steps. Additional ablation studies are provided in Appendix~\ref{appendix:additional_results}.

\label{subsec:abl}

\section{Conclusion}
\label{sec:conclusion}
In this work, we propose \ourmodel\ for degradation-aware optical flow, which estimates dense correspondences directly from corrupted inputs. Our approach starts from a pretrained image restoration diffusion model and extends it to multi-frame processing via spatio-temporal attention. This enables \ourmodel\ to leverage intermediate representations that encode both degradation-aware priors and geometric correspondence cues, while preserving spatial structure crucial for dense matching.
Extensive experiments on degraded optical flow benchmarks demonstrate that \ourmodel\ consistently improves flow estimation accuracy over existing methods.

\label{sec:conclusion}


\clearpage
\bibliographystyle{splncs04}
\bibliography{main}

\appendix
\clearpage
\setcounter{page}{1}
\renewcommand{\theHsection}{\Alph{section}}

\section*{\Large Appendix}
In this appendix, we provide additional details and experiments that supplement the main paper. Sec.~\ref{appendix:imple_details} describes the implementation details of \ourmodel. Sec.~\ref{appendix:feature_analysis} extends the feature analysis presented in Sec.~\ref{subsec:analysis} with further analysis and experiments. Sec.~\ref{appendix:additional_results} presents additional experiments and ablation studies. Sec.~\ref{appendix:quals} provides more qualitative results. Finally, Sec.~\ref{appendix:limit} discusses limitations and future work.

\section{Implementation Details}
\label{appendix:imple_details}
This section describes the implementation details of our experiments. Unless specified, all settings follow the default configurations.

\subsubsection{Lifting image restoration diffusion model.}
We adopt DiT4SR~\cite{duan2025dit4sr} as our image restoration diffusion model, using the official code with the publicly available \texttt{dit4sr\_q} weights, as in the original paper. To lift the model, we modify the attention operation at each layer by reshaping the frames, originally processed as a batch, into the spatial token dimension, enabling full attention to be computed across all frames within a single forward pass. We use the YouHQ~\cite{zhou2024upscale} dataset for training, and adopt the degradation pipeline from the STAR~\cite{xie2025star} codebase to synthesize low-quality frames as model inputs.

\subsubsection{Optical flow network.}
Our optical flow estimation network $\mathcal{M}_\phi$ is based on RAFT~\cite{teed2020raft}, initialized from the \texttt{raft-things.pth} weights pretrained on FlyingThings3D. For the feature upsampler, we adopt the DPT architecture following the VGGT~\cite{vggt} codebase. When fusing multi-scale features in DPT, we construct a feature pyramid with resolution scales of $(1, 1, 2, 2)$, where each value indicates the scaling factor relative to the input resolution for the corresponding pyramid level.

\subsubsection{Pseudo ground-truth for optical flow.}
To generate pseudo ground-truth optical flow for feature analysis in Sec.~\ref{subsec:analysis}, we apply SEA-RAFT~\cite{wang2024sea} on the high-quality videos using the official implementation with the \texttt{spring-M} configuration  (4 recurrent iterations) and the \texttt{Tartan-C-T-TSKH-spring540x960-M.pth} weights. For training \ourmodel, we use the same configuration with 20 recurrent iterations to obtain higher-quality pseudo ground-truth flow.

\subsubsection{Training and inference details.}
Unless specified in Sec.~\ref{subsec:exp_setup}, we follow the training settings of DiT4SR~\cite{duan2025dit4sr}. When training the lifting model, we generate text prompts using the captioner provided by the original model and use them as conditions, enabling the model to effectively learn cross-frame correspondences. For diffusion feature analysis, as well as training and inference of \ourmodel, we use null prompts for all inputs, as the low-quality input frames make it difficult to generate reliable captions. During inference, we use 10 denoising steps and follow the same procedure as training: the input video is processed in chunks of 3 frames through the model, and the resulting features are used to estimate optical flow with 20 refinement iterations.

\section{Additional Feature Analysis}
\label{appendix:feature_analysis}

\subsection{Full Feature Analysis Results}
\begin{figure}[h]
  \centering
  \includegraphics[width=1\linewidth]{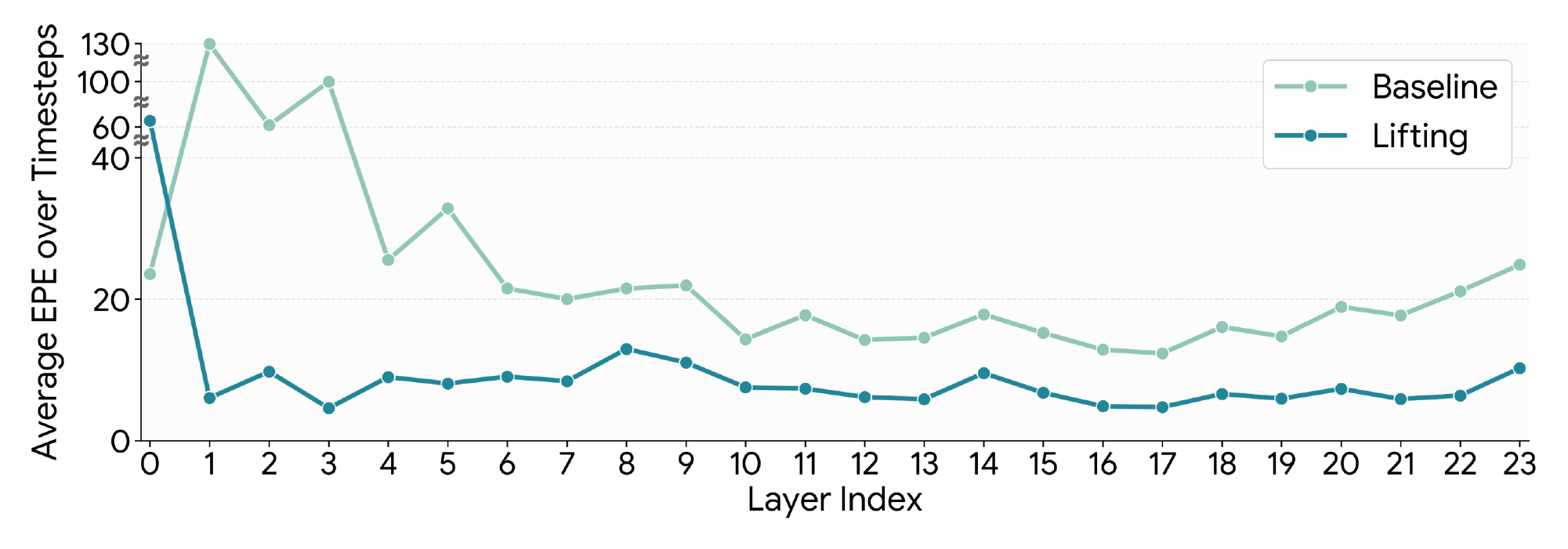}
  \caption{\textbf{Comparison of layer-wise average EPE over timesteps.}}
  \label{sup:figs:all_plot}
  \vspace{-10pt}
\end{figure}

Fig.~\ref{fig:sub1} presents the results for the top-10 features of both the baseline and lifting models. Extending this analysis, Fig.~\ref{sup:figs:all_plot} shows the timestep-averaged EPE across all layers for the baseline and lifting models. Across nearly all feature indices, the lifted features exhibit substantially lower EPE. This demonstrates that lifting enhances correspondence capability by enabling the model to learn cross-frame information.

\subsection{Comparison of Alternative Feature Type}
\begin{figure}[h]
    \centering
        \begin{subfigure}[t]{0.48\linewidth}
            \centering
            \includegraphics[width=\linewidth]{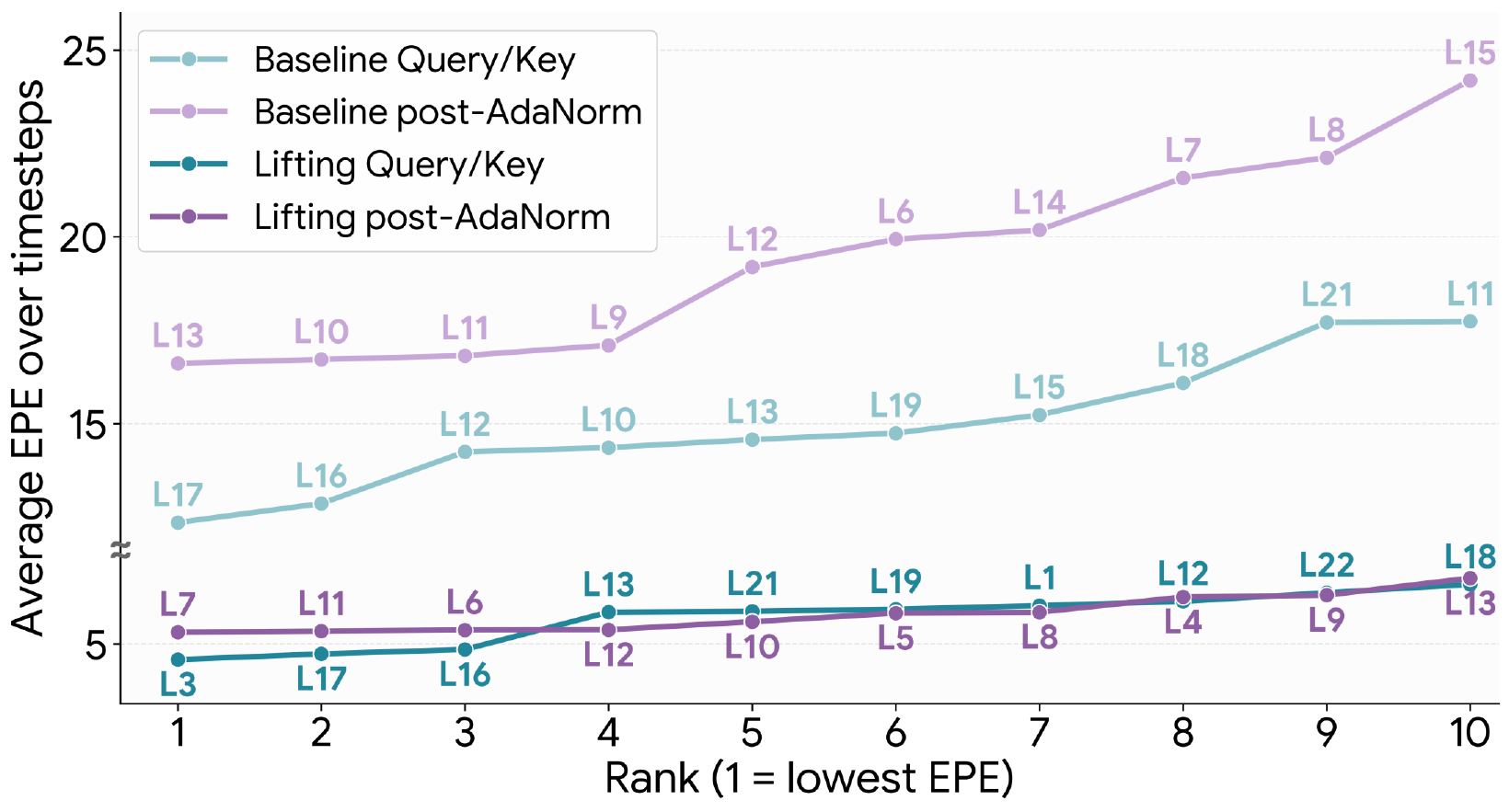}
            \caption{Top-10 layers by average EPE}
            \label{sup:fig:adanorm-1}
    \end{subfigure}
    \hfill
    \begin{subfigure}[t]{0.48\linewidth}
        \centering
        \includegraphics[width=\linewidth]{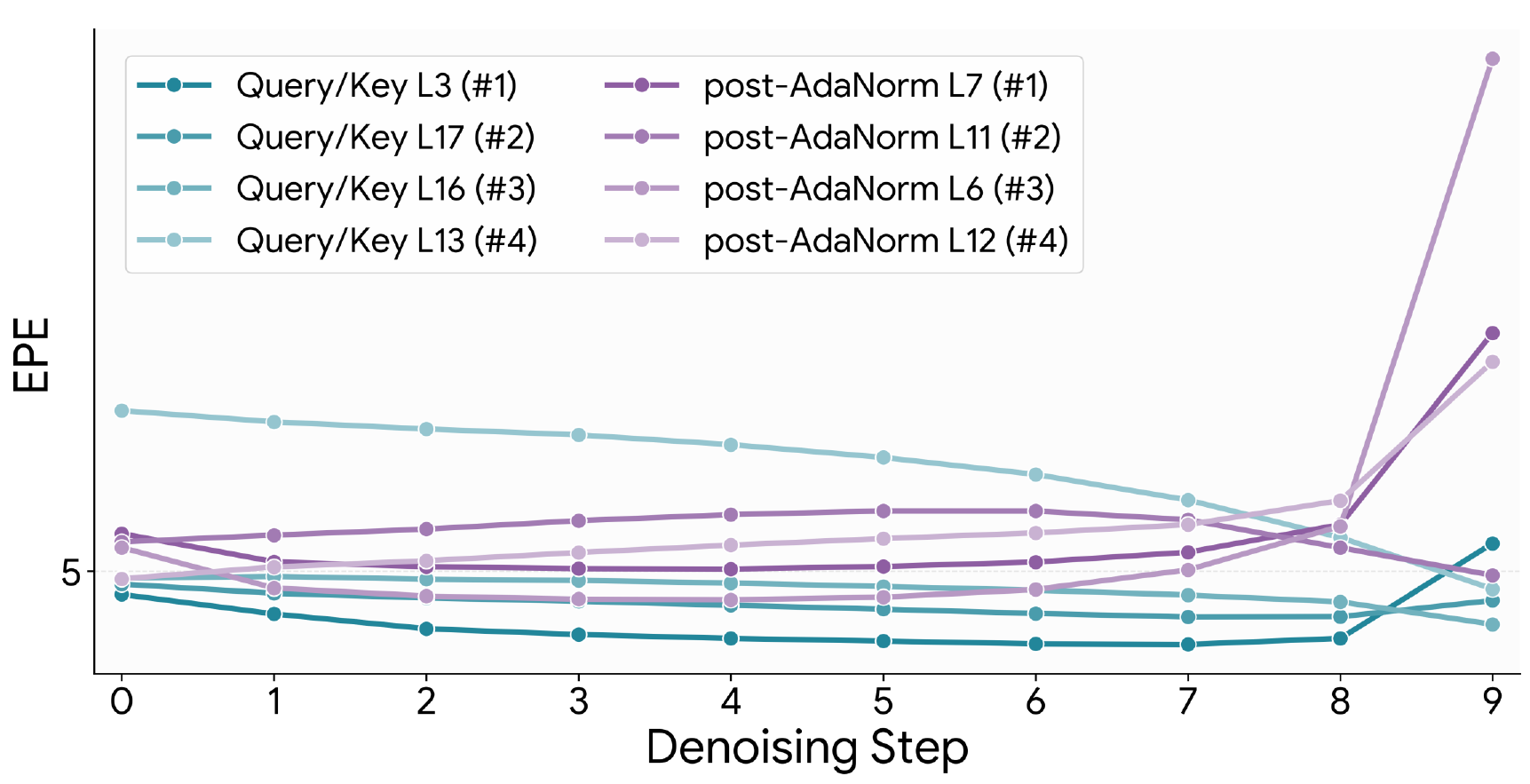}
        \caption{Top-4 Layer EPE over Denoising Steps}
        \label{sup:fig:adanorm-2}
    \end{subfigure}

    \makebox[\linewidth][c]{%
        \begin{subfigure}[t]{0.48\linewidth}
            \centering
            \includegraphics[width=\linewidth]{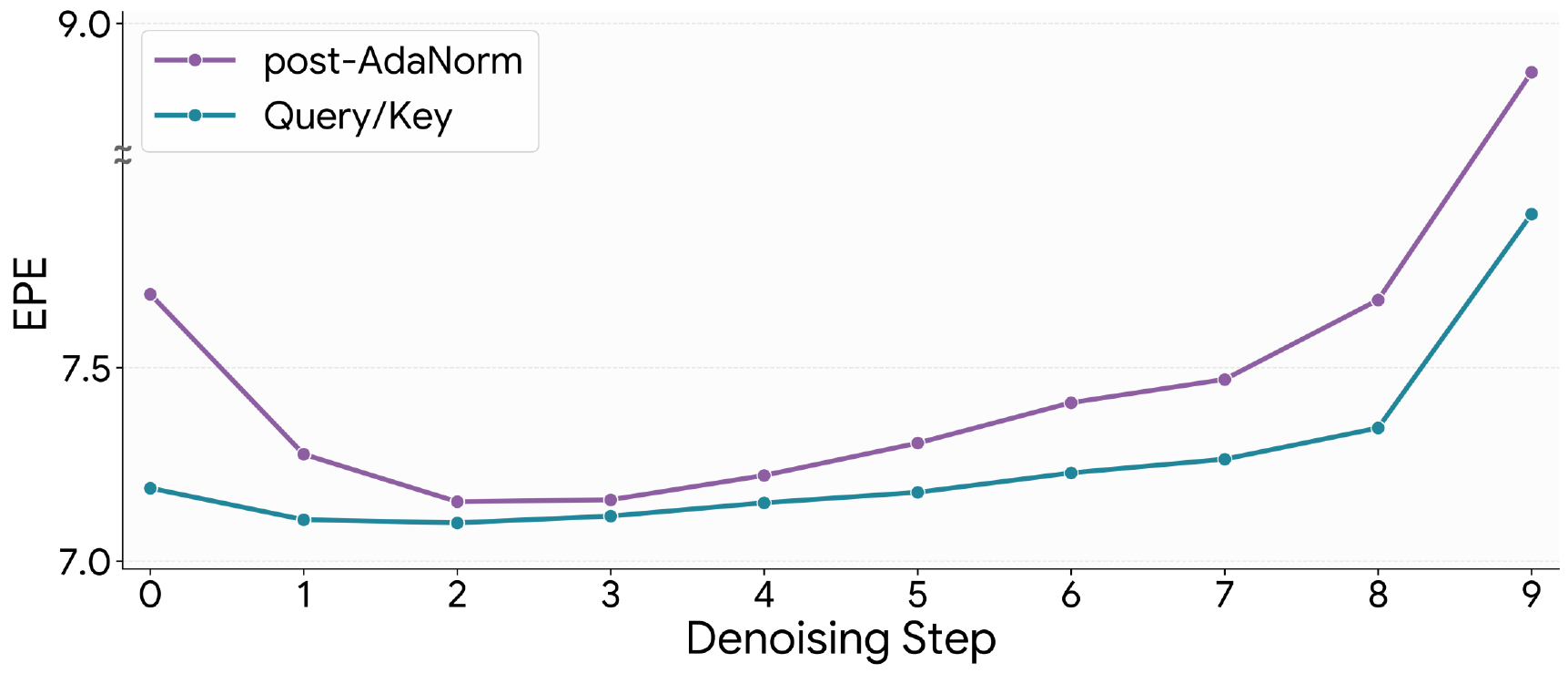}
            \caption{EPE over denoising steps after training}
            \label{sup:fig:adanorm-3}
        \end{subfigure}
    }
    \caption{\textbf{Comparison of zero-shot geometric correspondence between post-AdaNorm and Query/Key features.} (a) Top-10 layers ranked by timestep-averaged EPE (lower is better). (b) EPE over denoising steps for the top-4 layers of each method. (c) EPE on Sintel~\cite{butler2012naturalistic} over denoising steps after training the flow network. \ourmodel\ uses Query/Key features.}
    \label{sup:fig:adanorm}
\end{figure}
\ourmodel\ utilizes query and key features from the full attention layers of the diffusion model. However, features can also be extracted from other locations within each layer. Among the various candidates, we select post-AdaNorm features for comparison, as DITF~\cite{gan2025unleashing_DITF} has already demonstrated their effectiveness for semantic correspondence, making them a reasonable alternative to examine.

We extract post-AdaNorm features and conduct the same feature analysis described in Sec.~\ref{subsec:analysis}. Fig.~\ref{sup:fig:adanorm-1} ranks the top-10 layers by timestep-averaged EPE for both feature types. At the baseline level, query and key features consistently achieve lower EPE than post-AdaNorm features. After lifting, the gap narrows and the two become comparable except for the top-3 layers, where query and key features retain a clear advantage. To further investigate, we plot the EPE over denoising steps for the top-4 layers of each method in Fig.~\ref{sup:fig:adanorm-2}. Query and key features tend to show lower EPE overall, while post-AdaNorm features exhibit a noticeable spike at the final denoising step. Since the two feature types yield relatively similar trends, we further verify their difference by training the flow network using post-AdaNorm features in place of query and key features, keeping all other settings identical to \ourmodel. Fig.~\ref{sup:fig:adanorm-3} reports the resulting EPE on Sintel~\cite{butler2012naturalistic} over denoising steps, confirming that query and key features lead to better flow estimation after training as well. We conjecture that this advantage stems from the attention mechanism, which inherently encodes pairwise spatial relationships in the query and key projections, making them better suited for geometric correspondence.

\subsection{Video Restoration Diffusion Model}

\begin{figure}[h]
  \centering
  \includegraphics[width=0.7\linewidth]{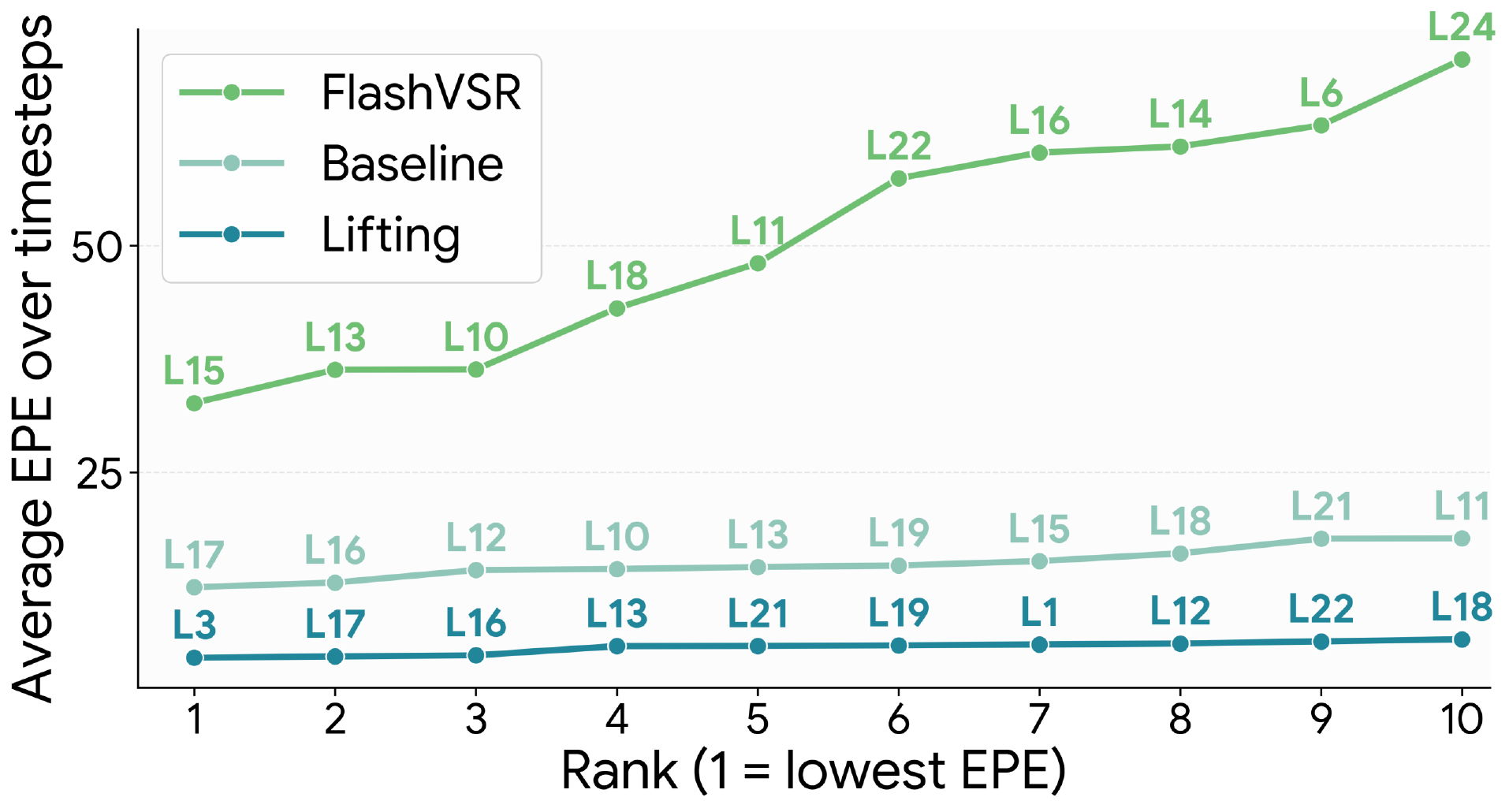}
  \caption{\textbf{Zero-shot geometric correspondence of Query/Key features from FlashVSR~\cite{zhuang2025flashvsr}.}}
  \label{sup:figs:flashvsr_feat}
\end{figure}
Video restoration diffusion models typically compress multiple frames into a single latent, making them inherently unsuitable for extracting frame-by-frame features. Nevertheless, we can still extract query and key features from such models and, analogous to how their VAE decodes a single latent into multiple frames, interpolate a single feature map to obtain per-frame features for the same analysis described in Sec.~\ref{subsec:analysis}. Fig.~\ref{sup:figs:flashvsr_feat} presents the results on FlashVSR~\cite{zhuang2025flashvsr}. Due to the interpolation of compressed features, geometric correspondence is significantly inferior to both the baseline and lifting, indicating that features from image diffusion models are more suitable for degradation-aware optical flow than those from video diffusion models.

\section{Additional Experiments}
\label{appendix:additional_results}

\subsection{Metrics over Denoising Steps}
Tab.~\ref{tab:main_quan} reports the quantitative results of \ourmodel\ averaged over all denoising timesteps. The full per-timestep results are presented in Tab.~\ref{sup:tab:plot_all_quan}. On Sintel and Spring, \ourmodel\ consistently outperforms existing methods across all timesteps. On TartanAir, while the timestep-averaged EPE is slightly higher than those of prior methods, \ourmodel\ achieves superior performance at both the initial (step 0) and final (step 9) denoising steps.

\begin{table*}[h]
  \centering
  \caption{\textbf{Timestep-wise quantitative results of \ourmodel\ on Sintel~\cite{butler2012naturalistic}, Spring~\cite{Mehl2023_Spring}, and TartanAir~\cite{wang2020tartanair}.} \colorbox{best}{Best} and \colorbox{secondbest}{second best} results are highlighted.}
  \label{sup:tab:plot_all_quan}
  \scriptsize
  \setlength{\tabcolsep}{3.6pt}
  \resizebox{\textwidth}{!}{%
  \begin{tabular}{ll*{14}{c}}
    \toprule
    & & \multicolumn{11}{c}{Step-wise Results of \ourmodel} & \multicolumn{3}{c}{Method} \\
    \cmidrule(lr){3-13}\cmidrule(lr){14-16}
    Dataset & Metric & 0 & 1 & 2 & 3 & 4 & 5 & 6 & 7 & 8 & 9 & Average & RAFT~\cite{teed2020raft} & SEA-RAFT~\cite{wang2024sea} & FlowSeek~\cite{poggi2025flowseek} \\
    \midrule
    \multirow{4}{*}{Sintel~\cite{butler2012naturalistic}}
      & EPE$\downarrow$  & 7.021 & 6.781 & \cellcolor{secondbest}6.720 & \cellcolor{best}6.716 & 6.743 & 6.761 & 6.803 & 6.874 & 7.064 & 7.640 & 6.912 & 10.69 & 10.18 & 10.24 \\
      & 1px$\downarrow$  & 57.04 & 56.81 & 56.39 & 55.97 & 55.52 & 55.12 & 54.79 & \cellcolor{secondbest}54.72 & \cellcolor{best}54.64 & 57.03 & 55.80 & 62.91 & 59.56 & 64.08 \\
      & 3px$\downarrow$  & 28.58 & 28.06 & 27.67 & \cellcolor{best}27.52 & \cellcolor{secondbest}27.56 & 27.71 & 27.88 & 28.14 & 28.40 & 29.52 & 28.10 & 37.24 & 34.46 & 40.71 \\
      & 5px$\downarrow$  & 21.33 & 20.80 & 20.49 & \cellcolor{best}20.39 & \cellcolor{secondbest}20.43 & 20.57 & 20.80 & 21.01 & 21.29 & 21.98 & 20.91 & 28.63 & 26.15 & 31.83 \\
    \midrule
    \multirow{4}{*}{Spring~\cite{Mehl2023_Spring}}
      & EPE$\downarrow$  & 2.290 & 2.212 & 2.207 & 2.203 & 2.201 & 2.204 & 2.201 & 2.193 & \cellcolor{secondbest}2.192 & \cellcolor{best}2.172 & 2.207 & 3.944 & 2.703 & 2.861 \\
      & 1px$\downarrow$  & \cellcolor{best}30.49 & 31.01 & 31.06 & 30.99 & 31.05 & 31.23 & 31.23 & 31.06 & 30.88 & \cellcolor{secondbest}30.54 & 30.95 & 39.82 & 41.51 & 41.53 \\
      & 3px$\downarrow$  & 14.16 & 13.95 & 13.94 & 13.94 & 13.95 & 13.96 & 13.91 & 13.82 & \cellcolor{secondbest}13.75 & \cellcolor{best}13.33 & 13.87 & 18.65 & 19.31 & 19.16 \\
      & 5px$\downarrow$  & {9.230} & {9.000} & {8.990} & {8.97} & {8.96} & \cellcolor{secondbest}{8.940} & \cellcolor{best}{8.910} & {8.940} & {8.940} & {8.940} & \cellcolor{best}{8.910} & 11.98 & 12.11 & 12.18 \\
    \midrule
    \multirow{4}{*}{TartanAir~\cite{wang2020tartanair}}
      & EPE$\downarrow$  & \cellcolor{best}6.674 & 8.675 & 8.926 & 8.955 & 9.076 & 9.384 & 9.696 & 10.06 & 9.951 & \cellcolor{secondbest}7.257 & 8.866 & 9.487 & 8.316 & 7.694 \\
      & 1px$\downarrow$  & 72.96 & 73.07 & 72.66 & 72.29 & 72.01 & 71.82 & \cellcolor{secondbest}71.69 & \cellcolor{secondbest}71.69 & \cellcolor{best}71.65 & 73.63 & 72.35 & 75.17 & 77.85 & 76.96 \\
      & 3px$\downarrow$  & \cellcolor{best}37.34 & 38.00 & 37.71 & 37.46 & \cellcolor{secondbest}37.38 & 37.42 & 37.47 & 37.53 & 37.47 & 38.28 & 37.61 & 42.96 & 45.76 & 45.20 \\
      & 5px$\downarrow$  & \cellcolor{best}24.57 & 25.53 & 25.34 & \cellcolor{secondbest}25.19 & 25.23 & 25.42 & 25.62 & 25.79 & 25.71 & 25.59 & 25.40 & 30.04 & 32.15 & 32.00 \\
    \bottomrule
  \end{tabular}%
  }
\end{table*}

\subsection{Finetuning RAFT}
\begin{table}[h]
  \centering
  \small
  \caption{\textbf{Average comparison on Sintel between RAFT* and \ourmodel.} Lower is better for all metrics.}
  \label{sub:tab:retrain_raft}
  \begin{tabular}{lcccc}
    \toprule
     & \multicolumn{4}{c}{Metric} \\
    \cmidrule(lr){2-5}
    Method & EPE $\downarrow$ & 1px $\downarrow$ & 3px $\downarrow$ & 5px $\downarrow$ \\
    \midrule
    RAFT* & 7.033 & 56.99 & 28.39 & \textbf{20.70} \\
    \ourmodel\ & \textbf{6.912} & \textbf{55.80} & \textbf{28.10} & 20.91 \\
    \bottomrule
  \end{tabular}
  \vspace{-12pt}
\end{table}
A straightforward baseline for our degradation-aware optical flow task is to finetune an existing flow network on the same training data. Specifically, we finetune RAFT~\cite{teed2020raft} using the identical setup as \ourmodel: low-quality and high-quality frame pairs generated by applying degradation kernels to YouHQ~\cite{zhou2024upscale}, with pseudo ground-truth flow obtained from SEA-RAFT~\cite{wang2024sea}. Tab.~\ref{sub:tab:retrain_raft} compares the finetuned RAFT~\cite{teed2020raft} (denoted RAFT*) with \ourmodel\ on Sintel. \ourmodel\ outperforms RAFT* in EPE, 1px, and 3px metrics, demonstrating that our approach offers clear advantages over simply finetuning a conventional flow network on the same data.

\subsection{Additional Ablation Studies}

\begin{table}[h]
\centering
\small
\setlength{\tabcolsep}{3pt}
\caption{\textbf{Ablation study on the architectural components of \ourmodel.} Row (d) corresponds to \ourmodel. \textbf{Bold} indicates the best result. Lower is better for all metrics.}
\label{sup:tab:arch_abl}
    \begin{tabular}{c c c|cccc}
    \toprule
    \multirow{2}{*}{} & \multicolumn{2}{c|}{Configuration} & \multicolumn{4}{c}{Metric} \\
    \cmidrule(lr){2-3} \cmidrule(lr){4-7}
     & Upsample & CNN Encoder & EPE $\downarrow$ & 1px $\downarrow$ & 3px $\downarrow$ & 5px $\downarrow$ \\
    \midrule
    (a) & Bilinear & \xmark & 7.2236 & 59.49 & 29.55 & 21.59 \\
    (b) & DPT & \xmark & 8.0745 & 64.67 & 30.07 & 21.95 \\
    (c) & Bilinear & \cmark & 7.1230 & 57.84 & 28.96 & 21.14 \\
    (d) & DPT & \cmark & \textbf{6.9122} & \textbf{55.80} & \textbf{28.10} & \textbf{20.91} \\
    \bottomrule
    \end{tabular}
\end{table}

In Sec.~\ref{sec:method}, we adopt DPT-based feature aggregation for upsampling and incorporate the CNN encoder from RAFT~\cite{teed2020raft} to better capture fine-grained local details beyond what diffusion features alone provide. To validate these design choices, we conduct an ablation study on the architectural components of \ourmodel, varying the upsampling strategy and the use of the CNN encoder. Tab.~\ref{sup:tab:arch_abl} reports EPE and pixel-threshold error rates (\%) on Sintel, averaged over all denoising timesteps, where row (d) corresponds to \ourmodel.

\subsubsection{CNN encoder.}
To assess the benefit of incorporating convolutional features from the RAFT~\cite{teed2020raft} encoder, we compare configurations with and without it under the same upsampling strategy. Adding the CNN encoder consistently improves performance in both the bilinear setting ((a) vs.\ (c)) and the DPT setting ((b) vs.\ (d)), with the latter showing particularly large gains. This confirms that the RAFT encoder provides complementary fine-grained spatial information that the diffusion features alone lack, and that the combination of DPT upsampling and CNN encoder features is essential for optimal flow estimation.

\subsubsection{Feature upsampling.}
We examine the effect of the upsampling strategy by comparing configurations with and without DPT-based feature aggregation. Without the CNN encoder, replacing bilinear interpolation with DPT ((a) vs.\ (b)) does not lead to consistent improvements across all metrics. However, when the CNN encoder is incorporated ((c) vs.\ (d)), DPT upsampling yields clear gains across all metrics including EPE. This suggests that DPT-based feature aggregation becomes effective when operating on sufficiently detailed features, and that the fine-grained information provided by the CNN encoder enables the DPT to fully leverage multi-scale aggregation for improved flow estimation.

\subsection{Application}
\begin{figure}[h]
    \centering
    \includegraphics[width=1.0\linewidth]{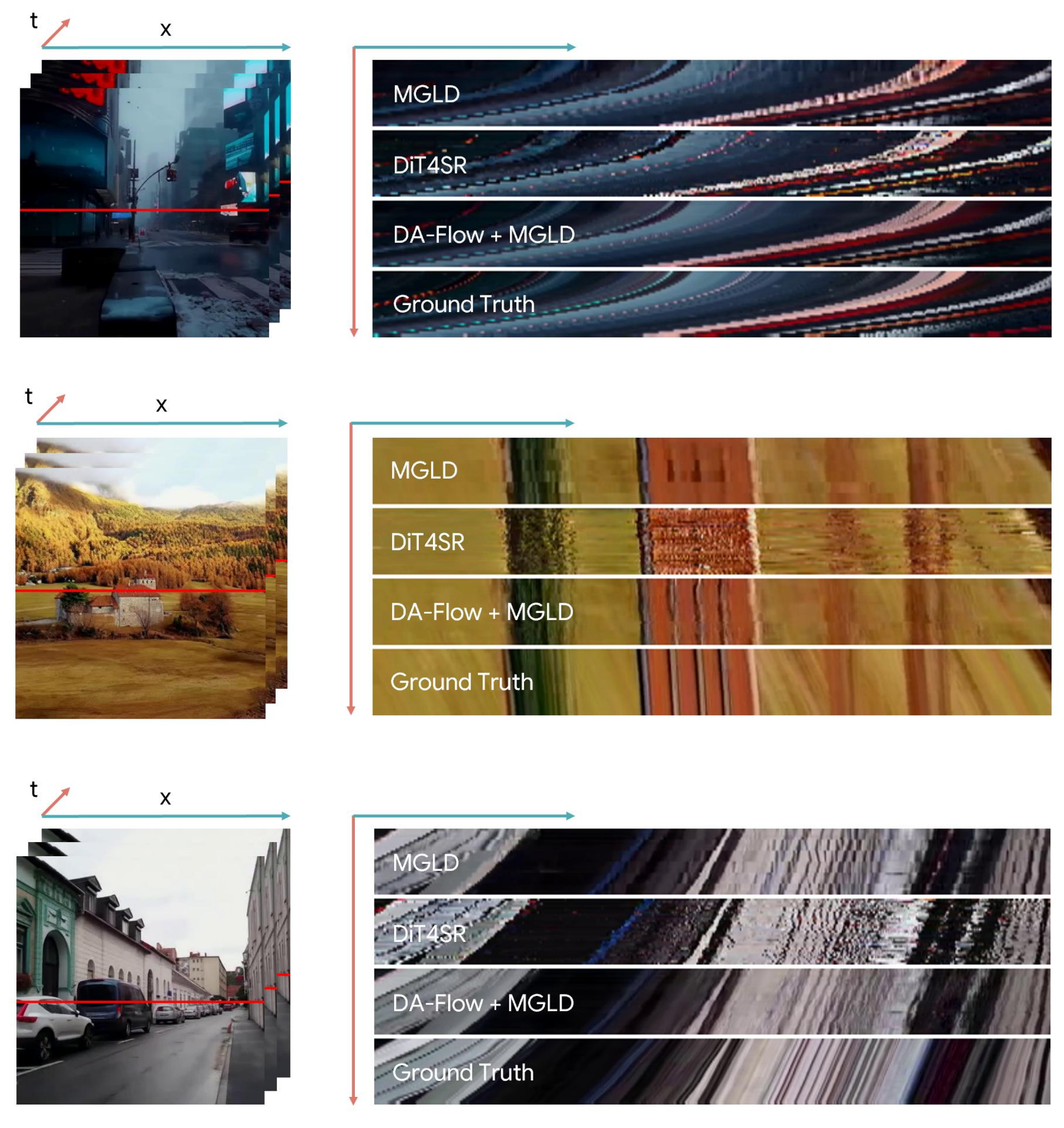} 
    \caption{
        \textbf{Comparison of temporal consistency in video restoration.} 
    }
    \label{sup:fig:qual_temporal}
\end{figure}

Several video restoration methods~\cite{zhou2024upscale, yang2024motion} leverage off-the-shelf optical flow models such as RAFT~\cite{teed2020raft} to align neighboring frames for improved temporal consistency. However, since these models directly take degraded frames as input, the estimated flow is often inaccurate, limiting the effectiveness of temporal alignment. Among them, MGLD~\cite{yang2024motion} employs a guidance mechanism that warps the current restored frame toward the next frame using estimated optical flow and minimizes the L2 distance between them as a guidance loss, enforcing temporal consistency during the diffusion sampling process. We adopt this guidance technique and combine it with our lifted image restoration diffusion model and \ourmodel. Unlike MGLD~\cite{yang2024motion}, which operates in latent space, we perform guidance in image space by decoding the latent at each step. 
We evaluate on the YouHQ40 dataset used in Sec.~\ref{subsec:analysis}, and additionally include frame-by-frame restoration results from our baseline DiT4SR~\cite{duan2025dit4sr} for comparison. As shown in Tab.~\ref{sup:tab:vr_quan}, our approach achieves strong performance in both reference-based metrics (PSNR, SSIM, LPIPS) and the non-reference metric DOVER, while attaining the best warping error, validating the effectiveness of the accurate flow estimated by \ourmodel. We provide qualitative comparisons of temporal consistency in Fig.~\ref{sup:fig:qual_temporal}. Compared to other methods, \ourmodel\ improves temporal alignment, reducing flickering and maintaining structural stability across consecutive frames.

\begin{table}[h]
    \centering
    \caption{\textbf{Video restoration results on YouHQ40.} Note that $E_\text{warp}^*=E_\text{warp} \times 10^{-3}$. \textbf{Bold} indicates the best result.}
    \resizebox{0.6\linewidth}{!}{%
    \begin{tabular}{l|ccccc}
        \toprule
        Method & PSNR$\uparrow$ & SSIM$\uparrow$ & LPIPS$\downarrow$ & DOVER$\uparrow$ & $E_\text{warp}^*\downarrow$\\
        \midrule
        MGLD         & {22.50} & 0.626 & 0.2661 & 47.41 & 4.532 \\
        DiT4SR       &  19.55 & 0.511 & 0.335 & \textbf{74.97} & {29.50} \\
        \ourmodel\ + MGLD & \textbf{23.47} & \textbf{0.646} & \textbf{0.215} & 57.96 & \textbf{3.483} \\
        \bottomrule
    \end{tabular}
    }
    \label{sup:tab:vr_quan}
\end{table}
\section{Additional Qualitative Results}
\label{appendix:quals}

\begin{figure}[t]
  \centering
  \includegraphics[width=1\linewidth]{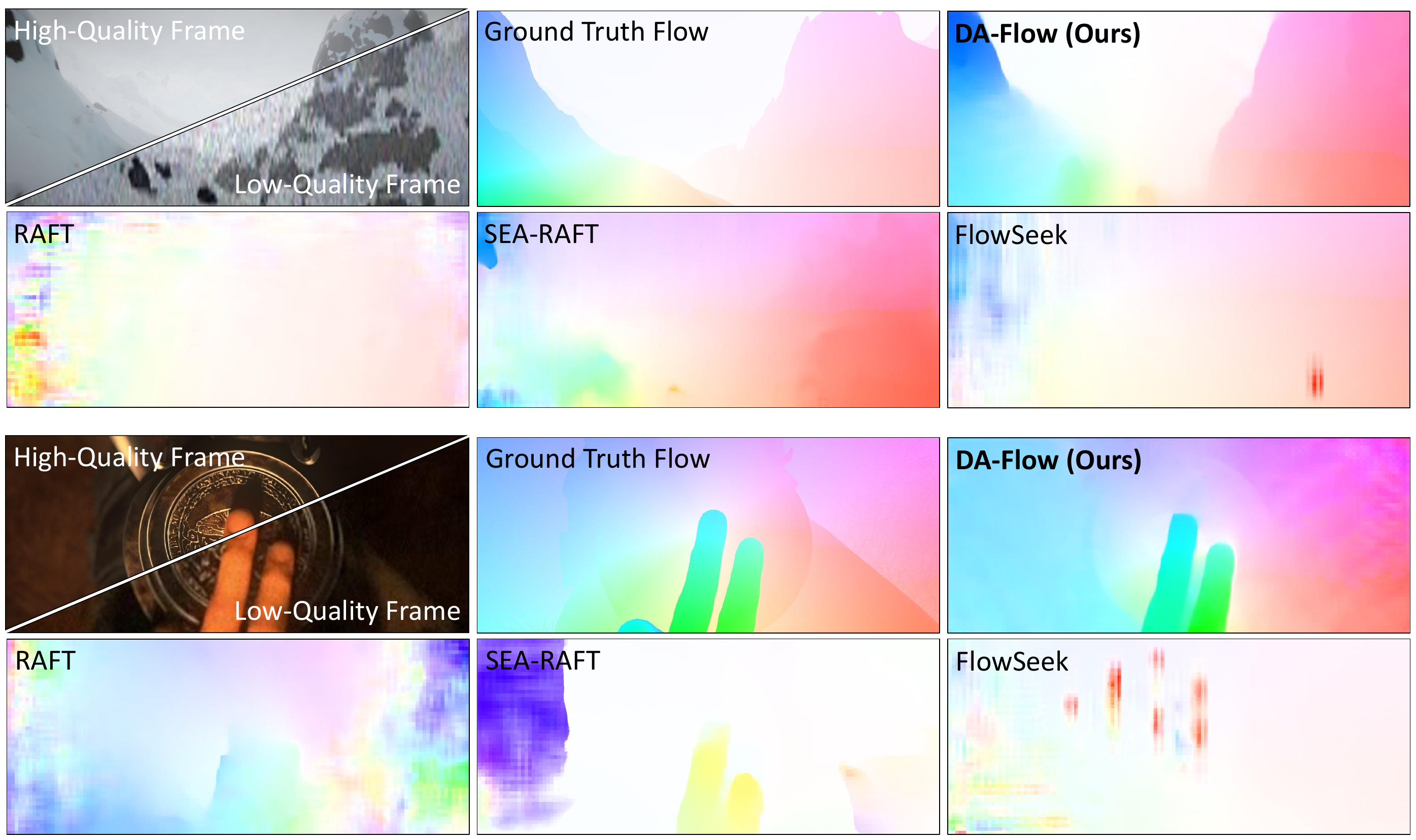}
  \caption{\textbf{Additional qualitative results on Sintel~\cite{butler2012naturalistic}.}}
  \label{figs:sup_qual_sintel}
  \vspace{-10pt}
\end{figure}

\begin{figure}[t]
  \centering
  \includegraphics[width=1\linewidth]{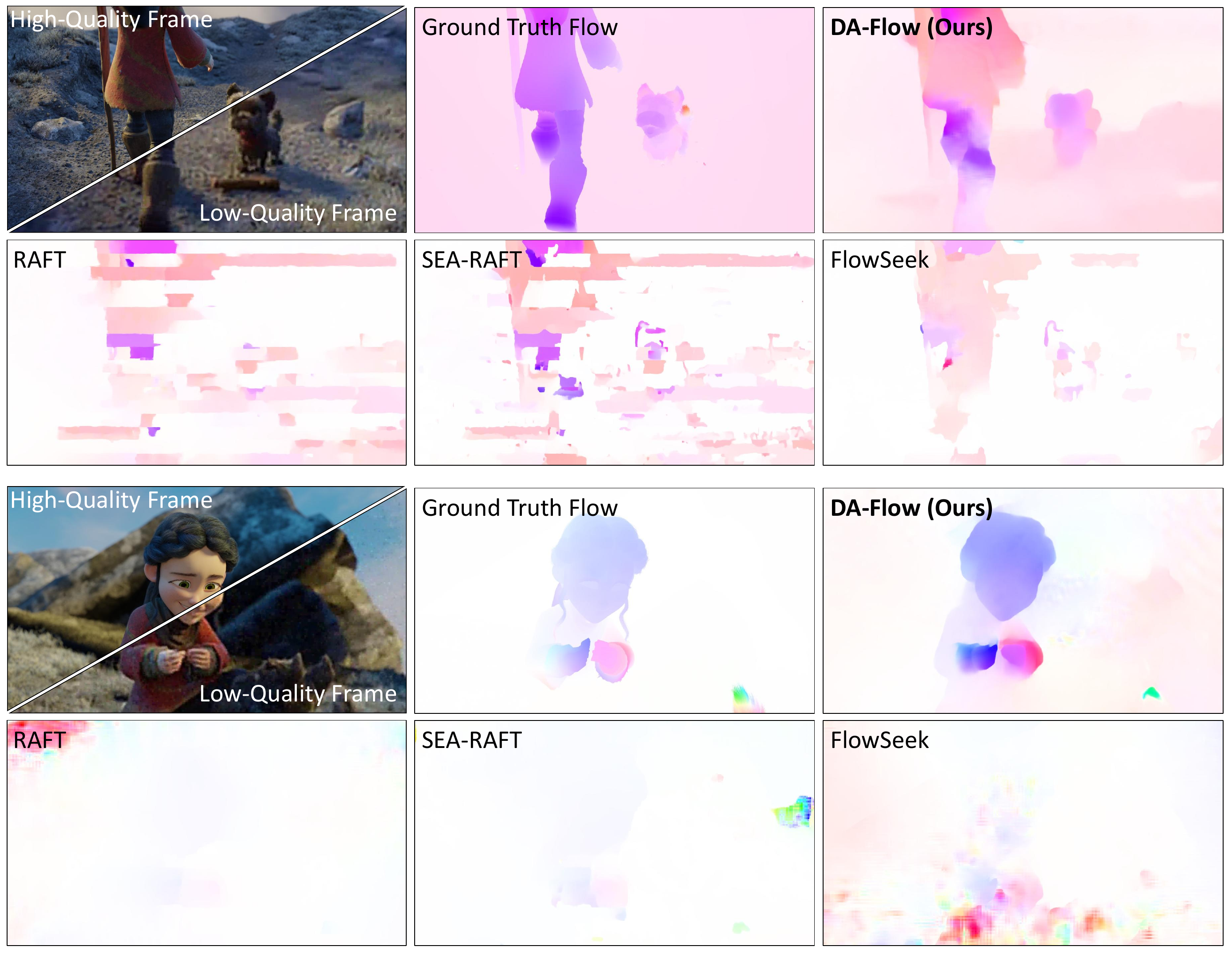}
  \caption{\textbf{Additional qualitative results on Spring~\cite{Mehl2023_Spring}.}}
  \label{figs:sup_qual_spring}
  \vspace{-10pt}
\end{figure}

\begin{figure}[t]
  \centering
  \includegraphics[width=1\linewidth]{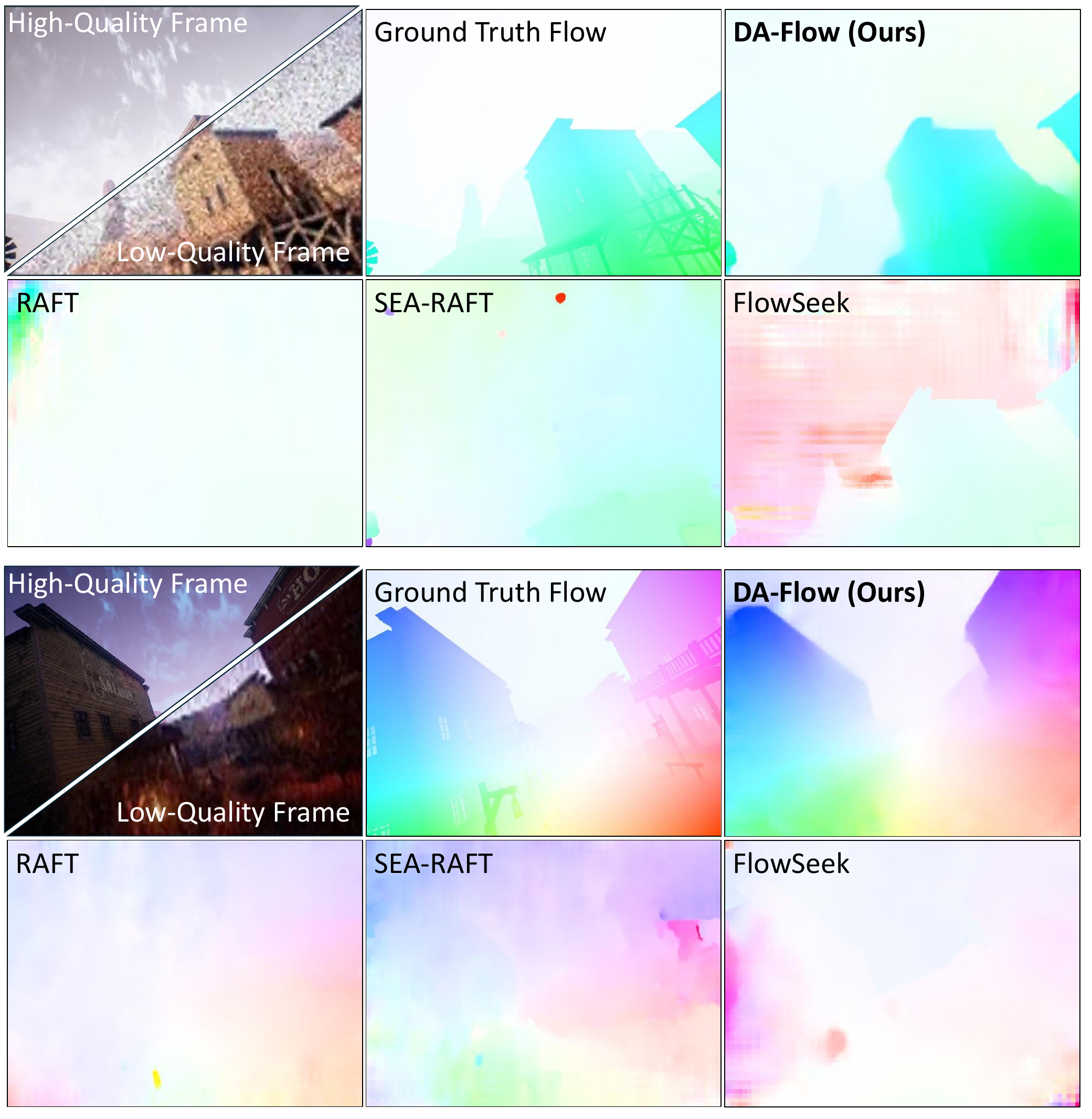}
  \caption{\textbf{Additional qualitative results on TartanAir~\cite{wang2020tartanair}.}}
  \label{figs:sup_qual_tartanair}
  \vspace{-10pt}
\end{figure}

We provide additional qualitative results on all benchmark datasets in  Fig.~\ref{figs:sup_qual_sintel}, Fig.~\ref{figs:sup_qual_spring}, and Fig.~\ref{figs:sup_qual_tartanair}. These examples further demonstrate the effectiveness of \ourmodel~across diverse scenarios.

\clearpage

\clearpage
\section{Limitations and Future Work}
\label{appendix:limit}

In this paper, we introduce degradation-aware optical flow, a new task that aims to accurately estimate flow from degraded video frames. Our approach leverages features from an image restoration diffusion model via lifting, which inherently requires multiple denoising steps at inference time, resulting in slower runtime compared to conventional flow estimation networks. Exploring one-step distillation techniques to reduce the inference cost while preserving estimation quality remains a promising direction for future work.
%
%
\end{document}